\documentclass[12pt]{article}

\usepackage{amssymb}
\usepackage{amsthm}
\usepackage{amsmath,bbm}
\usepackage{multirow}
\usepackage[table]{xcolor}
\usepackage{rotating}
\usepackage{pifont}
\usepackage{hhline,longtable,booktabs}
\usepackage{authblk}

\definecolor{gray}{gray}{0.8}

\title{Becoming More Robust to Label Noise with Classifier Diversity}

\author{Michael R. Smith\thanks{msmith@axon.cs.byu.edu}}
\author{Tony Martinez}
\affil{Department of Computer Science, Brigham Young University, Provo, UT 84602 USA}
\date{}
\begin{document}
\maketitle

\begin{abstract}
It is widely known in the machine learning community that class noise can be (and often is) detrimental to inducing a model of the data.
Many current approaches use a single, often biased, measurement to determine if an instance is noisy. 
A biased measure may work well on certain data sets, but it can also be less effective on a broader set of data sets \cite{Saez2013}.
In this paper, we present \textit{noise identification using classifier diversity} (\textit{NICD}) -- a method for deriving a less biased noise measurement and integrating it into the learning process.
To lessen the bias of the noise measure, NICD 
selects a diverse set of classifiers (based on their predictions of novel instances) to determine which instances are noisy.
We examine NICD as a technique for filtering, instance weighting, and selecting the base classifiers of a voting ensemble.
We compare NICD with several other noise handling techniques that do not consider classifier diversity on a set of 54 data sets and 5 learning algorithms.
NICD significantly increases the classification accuracy over the other considered approaches and is effective across a broad set of data sets and learning algorithms.
\end{abstract}

{\bf Keywords:} class noise; classifier diversity; filtering; instance weighting; machine learning; voting ensemble




\section{Introduction}
\label{section:intro}
The goal of supervised machine learning is to induce an accurate generalizing function from a set of labeled training instances.
However, most real-world data sets are noisy.
Generally, two types of noise are considered: attribute noise and label noise.
Previous work has found that, in general, label noise is more harmful than attribute noise \cite{Zhu2004,Nettleton2010}.
The consequences of label noise, as summarized by Fr\'{e}nay and Verleysen \cite{Frenay2014}, include 1) a deterioration of classification performance, 2) increased learning requirements and model complexity, and 3) a distortion of observed frequencies.
Knowing which instances are noisy and/or detrimental is non-trivial as, in most cases, all that is known about a task is contained in the set of training instances.

As discussed in the related work section, prior work has examined handling label noise using a variety of approaches that are generally specific to, or inspired by, an individual learning algorithm or information theoretic measure.
One commonly used approach removes the instances that are misclassified by a learning algorithm \cite{Tomek1976, John95}.
Although such an approach is biased towards the learning algorithm that is used, it has generally been shown to work well on the examined data sets and learning algorithms especially with the addition of artificial noise.
However, it has also been shown that the efficacy of a specific noise handling technique for a given learning algorithm is dependent upon the data set characteristics and, in some cases, using a noise handling technique \textit{reduces} the classification accuracy -- especially without the addition of artificial noise \cite{Saez2013, Smith_Eval}.
As ensembles often perform better than any one of its constituent base classifiers in classification \cite{Dietterich2000}, other prior work has used ensemble techniques to improve handling class noise \cite{Brodley1999,Verbaeten2003,Khoshgoftaar2004_ensembleFilter}. 
For an ensemble to be more accurate than any individual classifier of the ensemble, the base classifiers need to be accurate (better than random) and diverse \cite{Hansen1990}.
Using an ensemble technique for identifying noise implicitly lessens the dependence on a single hypothesis.
However, none of the previous work has explicitly focused on selecting diverse base classifiers when using an ensemble approach.

Inspired by the finding that noise handling is {\em not} always efficacious and using the principles of why ensembles increase classification accuracy (specifically diversity), we propose \textit{noise identification using classifier diversity} (NICD).
NICD first explicitly selects a set of diverse learning algorithms where diversity is determined by the predictions of the classifiers.
The diversity lessens the dependence of a noise measure on a specific hypothesis.
Without diversity, the same hypothesis could be over-represented if two hypotheses always classify the same way.
We examine using the set of diverse learning algorithms to 1) filter the instances, 2) weight the instances, and 3) as the base classifiers for a voting ensemble.

On a set of 54 data sets and 5 learning algorithms, we compare NICD with 8 filtering techniques, 2 weighting techniques, and a voting ensemble composed of different base classifiers -- all of which do not explicitly take classifier diversity into account.
We find that using classifier diversity significantly improves the accuracy for filtering, weighting, and voting ensembles across a \textit{broad} set of data sets, learning algorithms and noise levels -- demonstrating a robustness to label noise.
The term \textit{broad} refers to the characteristic that the noise handling method was {\em not} developed specifically for a given set of data sets and learning algorithms.
Overall, using NICD in a voting ensemble to select a diverse set of base classifiers achieves significantly higher classification accuracy then using a standard noise handling technique.

The remainder of the paper is organized as follows.
Section \ref{section:relatedWorks} reviews prior work in handling label noise. 
Section \ref{section:estimatingP} then presents our methodology for selecting diverse learning algorithms.
Our experimental methodology is presented in Section \ref{section:methodology}.
The results are provided in Section \ref{section:results}.
Section \ref{section:conclusions} concludes the paper.

\section{Related Work}
\label{section:relatedWorks}
As many real-wold data sets are inherently noisy, previous work has examined how class noise and attribute noise affects the performance of various learning algorithms \cite{Zhu2004,Nettleton2010}.
These works found that class noise is generally more harmful than attribute noise and that noise in the training set is more harmful that noise in the test set.
For a thorough overview of the current state of handling label noise in classification problems, consult the survey by Fr\'{e}nay and Verleysen \cite{Frenay2014}.

Most learning algorithms are designed to tolerate a certain degree of noise by making a trade-off between the complexity of the inferred model and minimizing error on the training data to prevent overfit.
For example, to avoid overfit many algorithms use a validation set for early stopping, pruning (such as in the C4.5 algorithm for decision trees \cite{Quinlan1993}), or regularization by adding a complexity penalty to the loss function \cite{bishop2006pattern}.
Despite these precautions, most learning algorithms are not completely robust to noise in the sense that the probability of an instance being misclassified changes with the presence of noise \cite{Frenay2014,Manwani2013}.

Some learning algorithms have been adapted specifically to better handle label noise.
For example, noisy instances are problematic for boosting algorithms \cite{Schapire1990,Freund1995} where more weight is placed upon misclassified instances--which often include mislabeled and noisy instances.
To address this, Servedio \cite{Servedio2003} presented a boosting algorithm that limits the amount of weight placed on any single training instance. 
In support vector machines (SVMs) with the standard hinge-loss function, misclassified instances (noisy instances are often misclassified) are always selected as support vectors.
To address this, Collobert et al. \cite{Collobert2006} use the ramp-loss function in a SVM to place a bound on the maximum penalty for an instance that lies on the wrong side of the margin.
Rosales et al. also modify a SVM to allow a certain number of instances to \textit{not} be considered as candidate support vectors \cite{Rosaloes2009}.
In a probabilistic setting, Lawrence and Sch\"{o}lkopf \cite{Lawrence2001} explicitly model the possibility that an instance is mislabeled using a generative model and then use expectation maximization to update the probability that an instance is mislabeled.

Preprocessing the data set is another approach that explicitly handles label noise.
This can be done by removing noisy instances, weighting the instances, or correcting incorrect labels.
All three approaches first attempt to identify which instances are noisy using various criteria.
Filtering noisy instances has received much attention and has generally been shown to result in an increase in classification accuracy, especially when large amounts of artificial noise are added to the data set \cite{Gamberger2000,Smith2011}.
However, recently the broad application of noise handling techniques has been shown to not always increase the classification accuracy, especially when no artificial noise has been added to the data set~\cite{Smith_Eval}.
Saez et al.~\cite{Saez2013} also showed that filtering can be detrimental and examined predicting the efficacy of filtering in nearest neighbor classification.
They found that there is some correlation between the data set properties and the efficacy of filtering.
One frequently used filtering technique removes any instance that is misclassified by a learning algorithm \cite{Wilson1972} or set of learning algorithms \cite{Brodley1999}.
Verbaeten and Van Assche \cite{Verbaeten2003} further pursued the idea of using an ensemble for filtering using ideas from boosting and bagging.
Other approaches use information theoretic or machine learning heuristics to remove noisy instances.
For example, Segata et al. \cite{Segata2009} remove instances that are too close or on the wrong side of the decision surface generated by a support vector machine.
Zeng and Martinez \cite{Xinchuan2003} remove instances that have a low probability of being labeled correctly where the probability is calculated using the output from a neural network.
Filtering has the potential downside of discarding useful instances.
However, it is assumed that there are significantly more non-noisy instances and that throwing away a few correct instances with the noisy instances will not have a negative impact on a large data set.

Weighting the instances in a training set has the benefit of not discarding any instances.
Filtering can be considered a special case of weighting where each instance is assigned a weight of 0 or 1.
In \textit{pair-wise expectation maximization}, Rebbapragada and Brodley \cite{Rebbapragada2007} weight the instances using expectation maximization to cluster instances that belong to a pair of classes.
The probabilities between classes for each instance is compiled and used to weight the influence of each instance.

Similar to weighting the training instances, data cleaning does not discard any instances, but rather strives to correct the noise in the instances.
As in filtering, the output from a learning algorithm has been used to clean the data.
Automatic data enhancement \cite{zeng.ida2001} uses the output from a neural network to correct the label for training instances that have a low probability of being correctly labeled.
Polishing \cite{Teng2000,Teng2003} trains a learning algorithm (in this case a decision tree) to predict the value for each attribute (including the class).
The predicted (i.e.~corrected) attribute values for the instances that increase generalization accuracy on a validation set are used instead of the uncleaned attribute values.

Most previous work has only considered the effects of a proposed noise handling technique on a small number of data sets and learning algorithms -- often only using a single learning algorithm.
Some studies have examined a larger set, but it is more of the exception rather than the rule \cite{Folleco2009,Khoshgoftaar2010}.
In this paper, we examine the effects of noise on a set of 54 data sets and 5 diverse learning algorithms.
NICD is differentiated from previous work in that NICD explicitly considers the diversity of the learning algorithms used to handle label noise to lessen the bias of using a single bias.

Prior work has also examined classifier diversity, primarily in the field of ensemble learning \cite{Brown2005_JMLR}.
Despite the efforts of prior work, a formal definition of diversity is still lacking \cite{Brown2005,Aksela2006} and the correlation between the classifier diversity measures and the classification accuracy of an ensemble method is often small \cite{Shipp2002, Kuncheva2003_ElusiveDiversity, Ruta2005}.
Most of the previous studies measured diversity only considering if a pair of classifiers both correctly classified or both misclassified an instance \cite{Kuncheva2003}.
To compute diversity, we use classifier output difference (COD) \cite{Peterson2005} which is distinguished from the other classifier diversity measures in that COD considers the case where the classifiers are both incorrect but predict different classes.
Some previous work has examined the effect of noise on the performance of ensembles, primarily examining the boosting, bagging, and decorate ensemble algorithms \cite{Dietterich2000_MLJ,McDonald2003,Melville2004} which create ensembles through subset selection of the training set.
To our knowledge, the impact of classifier diversity in the presence of class noise on a majority voting ensemble, or for noise identification, has not previously been examined.

\section{Identifying Noisy Instances with Classifier Diversity}
\label{section:estimatingP}
The first step in any noise handling technique is to identify noisy instances.
As discussed in Section \ref{section:relatedWorks}, a variety of approaches have been used to remove noisy instances, often using heuristics or biases from current learning algorithms or information theoretic measures.

In this paper, the probability $p(y|x)$ that an instance $\langle x,y\rangle$ will be correctly classified is used to determine how noisy each training instance is.
In practice, $p(y|x)$ is generally estimated using a specific hypothesis $h$ induced from a learning algorithm (i.e. $p(y|x)\approx p(y|x,h)$) -- which is a biased approximation.
The dependence of $p(y|x)$ on a specific $h$ can be removed by summing over all possible hypotheses $h$ in $\mathcal{H}$ and multiplying each $p(y|x,h)$ by the prior $p(h)$:
\begin{equation}
p(y|x) = \sum_{h\in\mathcal{H}} p(y|x,h)p(h). \label{eq:sumOverH}
\end{equation}
This formulation is infeasible, though, because 1) it is not practical (or possible) to sum over the set of all hypotheses, 2) calculating $p(h)$ is non-trivial, and 3) not all learning algorithms produce a probability distribution.
These limitations make probabilistic generative models attractive, such as the kernel Fisher discriminant algorithm \cite{Lawrence2001}.
However, for classification tasks, discriminative models generally have a lower asymptotic error than generative models \cite{Ng+Jordan:2001}.
In this paper, we choose to examine discriminative models.

\begin{figure}
\begin{center}
\input{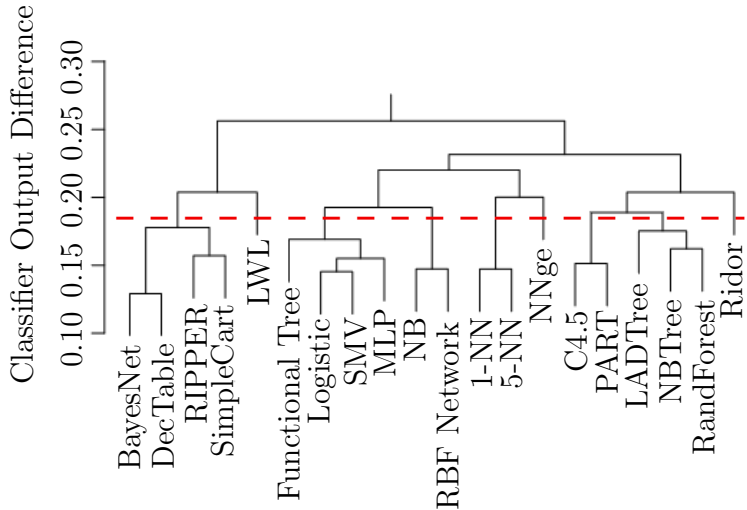}
\vskip 7mm
\caption{Dendrogram of the considered learning algorithms clustered using unsupervised metalearning based on their classifier output difference to calculate classifier diversity.}
\label{figure:COD}
\end{center}
\end{figure}

To lessen the dependence of $p(y|x)$ on a specific $h$, we estimate $p(y|x)$ using a diverse set of learning algorithms.
The diversity of the learning algorithms refers to the learning algorithms not having the same classification for all of the instances and is determined using unsupervised meta-learning (UML) \cite{Lee2011}.
UML first uses Classifier Output Difference (COD) \cite{Peterson2005} to measure the diversity between learning algorithms.
COD measures the distance between two learning algorithms as the probability that the learning algorithms make different predictions.
UML then clusters the learning algorithms based on their COD scores with hierarchical agglomerative clustering.
We considered 20 learning algorithms from Weka with their default parameters \cite{weka2009}.
The resulting dendrogram is shown in Figure \ref{figure:COD}, where the height of the line connecting two clusters corresponds to the distance (COD value) between them.
A cut-point of 0.18 was chosen to create 9 clusters and a representative algorithm from each cluster was chosen to create a diverse set of learning algorithms.
The set of learning algorithms $\mathcal{L}$ that are used to approximate $p(y|x)$ are listed in Table \ref{table:LA}.

\begin{table}
\caption{Set of diverse learning algorithms sampled from the clusters generated by UML.}
\label{table:LA}
\begin{center}
\setlength{\tabcolsep}{3pt}
\begin{tabular}{ll}
\hline
\multicolumn{2}{c}{Learning Algorithms}\\
\hline
*& Multilayer Perceptron trained with Back Propagation (MLP) \\
*& Decision Tree (C4.5) \cite{Quinlan1993} \\
*& Locally Weighted Learning (LWL) \\
*& 5-Nearest Neighbors (5-NN) \\
*& Nearest Neighbor with generalization (NNge) \\
*& Na\"{i}ve Bayes (NB) \\
*& RIpple DOwn Rule learner (RIDOR) \\
*& Random Forest (RandForest) \\
*& Repeated Incremental Pruning to Produce Error Reduction (RIPPER) \\
\end{tabular}
\end{center}
\end{table}

$p(y|x)$ is first approximated for each training instance using 10-fold cross-validation (the instance $\langle x, y \rangle$ is \textit{not} used to induce the model $h$).
Using the set of hypotheses induced by the diverse set of learning algorithms in $\mathcal{L}$, $p(y|x)$ is approximated as:
\begin{equation}
p(y|x) \approx p(y|x,\mathcal{L}) = \frac{1}{|\mathcal{L}|} \sum_{j=1}^{|\mathcal{L}|} p(y|x, l_j(T)) \label{eq:final}
\end{equation}
where $l_j(T)$ is the hypothesis from the $j^{th}$ learning algorithm trained on training set $T$.
From Equation \ref{eq:sumOverH}, $p(h)$ is estimated as $\frac{1}{|\mathcal{L}|}$ for the $j^{th}$ hypothesis generated from training the learning algorithms in $\mathcal{L}$ on $T$ and as zero for all of the other hypotheses in $\mathcal{H}$.

\section{Methodology}
\label{section:methodology}
In this section, we present our experimental methodology. 
\textit{Noise identification using classifier diversity} (NICD) is the process of selecting and using a diverse set of learning algorithms (Section \ref{section:estimatingP}) and then using this set of learning algorithms in the learning process to handle class noise.
In this paper, NICD uses the diverse set of learning algorithms for 1) filtering, 2) weighting, and 3) as the set of base classifiers for a voting ensemble.
We examine noise handling using the C4.5, 5-NN, MLP trained with backpropagation, random forest, and RIPPER learning algorithms on a set of 54 data sets from the UCI data repository \cite{UCI2010}.
Table \ref{table:dataSets} shows the data sets used in this study organized according to the number of instances, number of attributes, and attribute type.

\begin{table}
\centering
\caption{Datasets used organized by number of instances, number of attributes, and attribute type.}
\setlength{\tabcolsep}{5pt}
\begin{tabular}{| c | c | c | c | c |}
\hline
\multirow{2}{*}{\# Inst}& \multirow{2}{*}{\# Att} & \multicolumn{3}{| c |}{Attribute Type} \\
\cline{3-5} 
 &  & Categorical & Numerical & Mixed \\
\hline
\multirow{4}{*}{\begin{sideways}$M<100$ \end{sideways}} & \multirow{2}{*}{$k <$ 10} & Balloons & & Post-Operative \\
& & Contact Lenses & & \\
\cline{2-5}
& \multirow{1}{*}{$10 < k < 100$} & Lung Cancer & & Labor \\
\cline{2-5}
& \multirow{1}{*}{$k > 100$} & & Colon &  \\
\hline
\multirow{16}{*}{\begin{sideways}$100<M<1000$ \end{sideways}} & \multirow{6}{*}{$k < 10$} & Breast-w & Iris & Badges 2 \\
& & Breast Cancer & Ecoli & Teaching- \\
& & & Pima Indians & Assistant \\
& & & Glass & \\
& & & Bupa &  \\
& & & Balance Scale &  \\
\cline{2 - 5}
& \multirow{9}{*}{$10 < k < 100$} & Audiology & Ionosphere & Annealing \\
& & Soybean(large) & Wine & Dermatology \\
& & Lymphography & Sonar & Credit-A \\
& & Congressional- & Heart-Statlog & Credit-G \\
& & Voting Records & & Horse Colic \\
& & Vowel & & Heart-c \\
& & Primary-Tumor & & Hepatitis \\
& & Zoo & & Autos \\
& & & & Heart-h \\
\cline{2-5}
& \multirow{1}{*}{$k > 100$} & &  & Arrhythmia \\
\hline
\multirow{7}{*}{\begin{sideways}$ 1000 < M < 10000$ \end{sideways}} & \multirow{3}{*}{$k < 10$} & Car Evaluation & Yeast & Abalone \\
& & Chess & & \\
& & Titanic & & \\
\cline{2-5}
& \multirow{4}{*}{$k < 100$} & Splice & Waveform-5000 & Thyroid- \\
& &  & Segment & (sick \& \\
& &  & Spambase & hypothyroid) \\
& &  & Ozone &  \\
\hline
\multirow{6}{*}{\begin{sideways}$ M > 10000$ \end{sideways}} & \multirow{2}{*}{$k < 10$} & Nursery & MAGIC- &  \\
 & & & Telescope & \\
\cline{2-5}
 & \multirow{4}{*}{$k < 100$} & Chess- & & Adult-Census- \\
& & (King-Rook vs. &  & Income (KDD) \\
& &  King-Pawn) & & \\
& & Letter & & \\
\hline
\end{tabular}
\label{table:dataSets}
\end{table}

Each noise handling method is evaluated by averaging the results from ten runs of each experiment. 
For each experiment, the data is shuffled and then split into 2/3 for training and 1/3 for testing.
The training and testing sets are stratified.
Random noise is then introduced by randomly changing the class label of $n\%$ of the training instances where a new label is chosen uniformly from the possible class labels (noisy completely at random).
The random noise levels are examined at 0\%, 10\%, 20\%, 30\%, and 40\%.
Statistical significance between pairs of algorithms is determined using the Wilcoxon signed-ranks test as suggested by Dem\v{s}ar \cite{Demsar2006}.

We approximate $p(y|x)$ using classifier diversity (NICD) as well as a biased approach.
For NICD, the quantity $p(y|x)$ is approximated as $p(y|x,\mathcal{L})$ using all 9 learning algorithms listed in Table \ref{table:LA}.
Generally, classification learning algorithms classify an instance into nominal classes and do not produce a probability distribution over the classes. 
Since not all learning algorithms produce a probability distribution and to be consistent, we use the Kronecker delta function $\delta(h(x),y)$ instead of $p(y|x,l_j(T))$ (from Equation \ref{eq:final}) to produce a real-valued score for $p(y|x,\mathcal{L})$.

For the biased approach, $p(y|x)$ is approximated as $p(y|x,h)$ where the hypothesis $h$ is induced by the same learning algorithm that is used to induce a model of the data.
For biased filtering, misclassified instances are removed.
To get a real-value for biased instance weighting from a single hypothesis, we compute a classifier score for each instance from the learning algorithm that induces $h$.
Below, we present how we calculate the classifier scores for the investigated learning algorithms.
\begin{description}
 \item[{\bf Multilayer Perceptron (MLP):}]
For multiple classes, each class from a data set is represented with an output node.
After training a MLP with backpropagation, the classifier score is the largest value of the output nodes normalized between zero and one: 
$$
\hat{p}(y | x) = \frac{o_i(x)}{\sum_i^{|Y|} o_i(x)}
$$
where $y$ is a class from the set of possible classes $Y$ and $o_i$ is the value from the output node corresponding to class $y_i$.

 \item[{\bf Decision Tree:}]
To calculate a classifier score, an instance first follows the induced set of rules until it reaches a leaf node.
The classifier score is the number of training instances that have the same class as the examined instance divided by all of the training instances that also reach the same leaf node.

 \item[{\bf 5-NN:}]
5-NN returns the percentage of the nearest-neighbors that agree with the class label of an instance as the classifier score.

 \item[{\bf Random Forest:}]
Random forests return the class counts from the leaf nodes of each tree in the forest.
The counts for each class are summed together and then normalized between 0 and 1.

 \item[{\bf RIPPER:}]
RIPPER returns the percentage of training instances that are covered by a rule and share the same class as the examined instance.
\end{description}
Obviously, a classifier score does not produce a true probability.
However, the classifier scores approximate the confidence of an induced model for the class label of an instance.

For filtering using $p(y|x,\mathcal{L})$ (the $\mathcal{L}$-filter), instances that are misclassified by 50 percent of the learning algorithms in the ensemble are filtered from the training set. 
Note that other percentages could also be used.
We found that 50\% generally produces good results compared to values of 70\% and 90\%.
In practice a validation set is often used to determine the percentage that would be used.
For the biased filter approach (biased-filter), any instance that is misclassified by the same learning algorithm that is being used to induce a model of the data is filtered from the training set.

%


\begin{table}[t]
\centering
\caption{How instance weighting is integrated into the examined learning algorithms.}
\begin{tabular}{l|c|c}
Learn Alg & Orig & NICD\\
\hline
MLP & $(t-o)f'(net)$ & $p(y|x)(t-o)f'(net)$\\
\hline
Rand Forest & Uniform dist & Weighted by $p(y|x)$\\
\hline
C4.5, IB5,& Count number & Sum $p(y|x)$ \\
RIPPER & of instances, i.e. &\\ 
 & \multirow{2}{*}{$\frac{\sum_{c_i} 1}{\sum_T 1}$} & \multirow{2}{*}{$\frac{\sum_{c_i} p(y_i|x_i)}{\sum_T p(y_i|x_i)}$}\\
&  & \\
\end{tabular}
\label{table:QW}
\end{table}

Table \ref{table:QW} summarizes how an instance is weighted by (an approximation of) $p(y|x)$ for the examined learning algorithms. 
For MLPs trained with backpropagation, the error ($(t-o)f'(net)$) is scaled by $p(y|x)$ where $(t-o)$ is the difference between the target value and the output of the network, $f'(net)$ is the derivative of the activation function $f$ and $net$ is the sum of the product of each input $i_j$ and its corresponding weight $w_j$: $net=\sum_j w_ji_j$.
For Random Forests, the distribution for selecting instances in the random trees is weighted by $p(y|x)$ rather than being uniformly weighted. 
For the other learning algorithms that keep track of counts, each instance is weighted by $p(y|x)$.
$\sum_{c_i}$ represents summing over instances that meet some criterion $c_i$ and $\sum_T$ sums over all of the instances in the data set.

\section{Results}
\label{section:results}
In this section, we present the results of our experiments.
We compare $\mathcal{L}$-filtering with the biased-filter and seven other filtering approaches.
$\mathcal{L}$-weighting is compared with biased-weighting and one other instance weighting technique.
The compared noise handling techniques are listed below with a brief explanation.
For more details, consult the cited work.
\begin{description}
 \item [Repeated-edited nearest neighbor (RENN)~\cite{Tomek1976}.] 
RENN~repeatedly\linebreak removes the instances that are misclassified by a nearest neighbor classifier and has been shown to produce good results.
Here we set the number of nearest neighbors to 5.
 \item[Saturation filter~\cite{Gamberger2000}.] The saturation filter is based on the premise that removing noisy instances reduces the complexity of the least correct hypothesis (CLCH) while removing correctly labeled instances does not.
The instance whose removal from the training set decreases the CLCH the most is filtered from the training set.
This process continues until the CLCH can no longer be reduced or all of the training instances have been removed.
 \item[Classification filter~\cite{Gamberger1999}.] The classification filter removes instances that are misclassified by a learning algorithm using 10-fold cross-validation on the training set.
The default learning algorithm is a 1-nearest neighbor. 
 \item[Ensemble filter~\cite{Brodley1999}.] The ensemble filter removes instances that are misclassified by $n\%$ of the base classifiers.
The ensemble filter is distinguished from the $\mathcal{L}$-filter based on how the base classifiers are chosen for the ensemble.
For the ensemble filter, the authors arbitrarily chose three well-known learning algorithms (C4.5, IB1, and thermal linear machine \cite{Brodley1995}).
They then examined removing instances that were misclassified by the majority or all of the learning algorithms.
In this paper, instances are removed that are misclassified by all of the learning algorithms.
 \item[Automatic noise removal filter (ANR)~\cite{Xinchuan2003}.] ANR estimates the probability of each possible class for the training instances.
The probabilities are used during training such that an instance that has a low probability for its assigned class is ``corrected'' to the class that has the highest probability.
The instances that were corrected during training are then filtered from the data set.
 \item[Cross-validated committees filter~\cite{Verbaeten2003}.] The cross validated committees filter partitions a data set into $n$ subsets of approximately equal size and a learning algorithm is induced $n$ times, each time leaving out one of the subsets from the training data.
The $n$ classifiers are then ensembled together to determine if an instance is noisy.
Instances that are misclassified by the all of the ensembled classifiers are then filtered from the training set.
The base classifier used is a decision tree.
 \item[Iterative-partitioning filter~\cite{Khoshgoftaar2007}.] The iterative partitioning filter first\linebreak partitions the training data into $n$ subsets and a model is induced on each subset.
Instances that are misclassified by all of the induced models are filtered.
This process is repeated until the number of noisy instances that are removed is less than 1\% of the size of the original training set.
The base classifier used is a decision tree trained using C4.5.
 \item[Pair-wise expectation maximization (PWEM) \cite{Rebbapragada2007}.] PWEM weights\linebreak each instance using the EM algorithm.
First each data set is binarized.
For each pair of classes, the instances that belong to the two classes are clustered using EM where the number of clusters is determined using the Bayesian Information Criterion \cite{Kass1995}. 
Given the $Y-1$ clusterings ($Y$ is the number of classes in the data set), $p(y|x)$ is calculated as:
\begin{equation}
 p(y|x) = \sum_\theta p(\theta)p(y|x, \theta) = \sum_\theta p(\theta)\sum_{c=1}^k p(y|c_k,\theta)p(c_k|x,\theta)
\end{equation}
where $\theta$ is a clustering model induced using the EM algorithm, $c$ is a cluster in $\theta$ and $k$ is the number of clusters formed in $\theta$.
\end{description}
All of the previous techniques were used with their default parameters as implemented in the KEEL toolkit~\cite{KEEL} except RENN and PWEM.
In some cases, an algorithm did not finish running on all of the data sets.
For example, since the saturation filter iteratively removes one instance it requires large amounts of memory/time to run on large data sets and did not finish in some cases.
Rather then the remove the large data sets from the examination, we include the comparison on the data sets on which the algorithms did finish.

For the tables in this section, the algorithm in the first row is the baseline algorithm that the algorithms in the subsequent rows are compared against.
The values in the ``Count'' rows represent the number of times that the accuracy from the baseline algorithm is greater than, equal to, or less than the compared algorithm.
The bold \textit{p}-values represent the cases where the baseline algorithm achieves significantly higher classification accuracy.
In contrast, the underlined values in the gray cells represent the cases where the accuracy from the compared algorithm is significantly higher than the baseline algorithm.
A \textit{p}-value of 0.000 represents a \textit{p}-value that is less than 0.001.

\subsection{The Broad Application of Noise Handling with No Artificial Noise}

Many of the previous works in noise handling were inspired by and, often, biased towards a given learning algorithm or measure.
Most previous work added artificial noise to the data sets to evaluate their techniques and showed significant improvements on the data sets with artificial noise.
Due to the biased nature of the noise handling techniques, it is not surprising that the broad application of a noise handling approach to data sets without artificial noise can be detrimental \cite{Saez2013, Smith_Eval}.
We briefly re-examine the application of noise handling to a broad set of data sets and learning algorithms.
Table \ref{table:noNoise} compares no noise handling with the considered noise handling techniques averaged over all 54 data sets with {\em no artificially added noise}.
There are only 3 cases for which noise handling significantly increases the classification accuracy.
RENN, ANR, and the classification filter significantly \textit{decrease} the classification accuracy for all of the investigated learning algorithms.
This is an important example that highlights a point that is often overlooked in the noise handling literature -- noise handling can be detrimental if used in all cases.
Previous work has generally considered only a few data sets where noise handling is beneficial.

\begin{table}
\centering
\caption{The results of the 5 considered learning algorithms using the investigated noise handling approaches with no artificial noise added to the data sets averaged over the 54 data sets.
Bold \textit{p}-values represent cases where noise handling significantly {\em decreases} the accuracy; underlined, gray cells represent cases where noise handling significantly increases the accuracy.}
\setlength{\tabcolsep}{4.5pt}
\begin{tabular}{l|ccccc}
& C4.5 & IB5 & MLP & RF & RIP \\
\hline

Orig & 79.31 & 79.37 & 81.67 & 81.18 & 78.35 \\
\hline
NICD: $\mathcal{L}$-Weighted & 78.36 & 78.72 & 82.26 & 80.82 & 77.86 \\
\textit{p}-val & 0.475 & 0.162 & \cellcolor{gray}\underline{0.974} & 0.323 & 0.240 \\
Count & 26,1,27 & 27,4,23 & 18,3,33 & 28,2,24 & 26,2,26 \\
\hline
PWEM & 76.41 & 78.02 & 82.79 & 81.51 & 74.17 \\
\textit{p}-val & 0.062 & \textbf{0.003} & 0.737 & 0.125 & $\mathbf{<0.001}$ \\
Count & 30,3,21 & 33,3,18 & 23,3,28 & 34,1,19 & 39,1,14 \\
\hline
NICD: $\mathcal{L}$-Filter & 79.55 & 79.40 & 81.80 & 81.66 & 78.98 \\
\textit{p}-val & 0.913 & 0.854 & 0.584 & 0.873 & 0.893 \\
Count & 25,11,18 & 23,9,22 & 23,4,27 & 28,2,24 & 27,5,22 \\
\hline
RENN & 76.83 & 76.99 & 78.80 & 78.20 & 76.65 \\
\textit{p}-val & \textbf{0.002} & $\mathbf{<0.001}$ & $\mathbf{<0.001}$ & $\mathbf{<0.001}$ & \textbf{0.004} \\
Count & 32,3,19 & 35,4,15 & 38,1,15 & 35,2,17 & 34,2,18 \\
\hline
ANR & 67.25 & 67.64 & 69.17 & 68.20 & 68.18 \\
\textit{p}-val & $\mathbf{<0.001}$ & $\mathbf{<0.001}$ & $\mathbf{<0.001}$ & $\mathbf{<0.001}$ & $\mathbf{<0.001}$ \\
Count & 34,3,17 & 34,7,13 & 37,2,15 & 36,3,15 & 40,1,12 \\
\hline
ClassificationFilter & 73.65 & 73.09 & 75.65 & 75.11 & 72.48 \\
\textit{p}-val & \textbf{0.009} & $\mathbf{<0.001}$ & \textbf{0.002} & \textbf{0.012} & \textbf{0.001} \\
Count & 32,3,19 & 36,4,14 & 34,2,18 & 32,1,21 & 34,3,17 \\
\hline
CVCommitteesFilter & 79.41 & 79.37 & 82.33 & 81.72 & 78.83 \\
\textit{p}-val & 0.935 & 0.492 & 0.947 & \cellcolor{gray}\underline{0.958} & \cellcolor{gray}\underline{0.965} \\
Count & 18,10,26 & 22,14,18 & 20,4,30 & 21,2,31 & 16,8,30 \\
\hline
EnsembleFilter & 79.53 & 79.06 & 81.62 & 80.91 & 78.87 \\
\textit{p}-val & 0.940 & 0.140 & 0.504 & 0.253 & 0.875 \\
Count & 24,2,28 & 32,1,21 & 28,1,25 & 31,1,22 & 23,3,28 \\
\hline
IterPartitionFilter & 79.31 & 79.27 & 82.18 & 81.51 & 78.78 \\
\textit{p}-val & 0.782 & 0.465 & 0.862 & 0.777 & 0.744 \\
Count & 16,13,25 & 23,10,21 & 21,4,29 & 24,4,26 & 23,5,26 \\
\hline
SaturationFilter & 78.70 & 79.08 & 81.07 & 80.28 & 78.46 \\
\textit{p}-val & 0.418 & 0.206 & 0.216 & 0.089 & 0.402 \\
Count & 21,9,20 & 24,9,18 & 26,4,20 & 29,2,19 & 22,7,21 \\

\end{tabular}
\label{table:noNoise}
\end{table}

\begin{figure}
\centering
\setlength{\tabcolsep}{-11pt}
\renewcommand{\arraystretch}{-1}
\begin{tabular}{cc}
\input{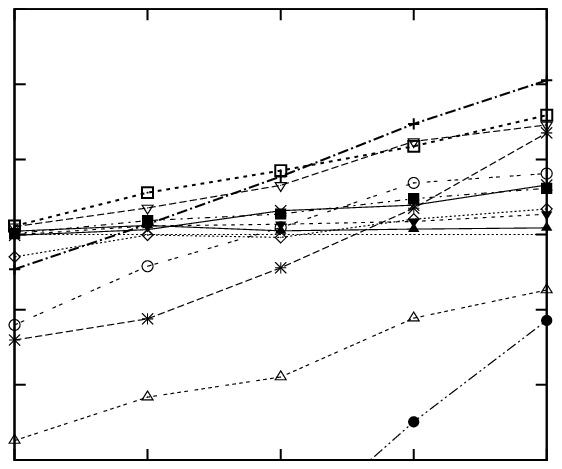} & \input{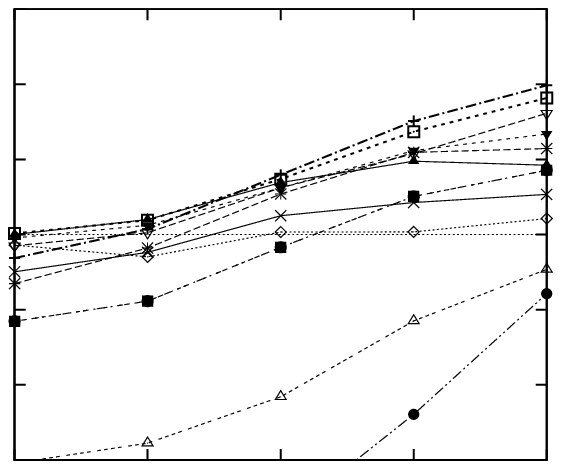}\\
C4.5 & IB5\\
\input{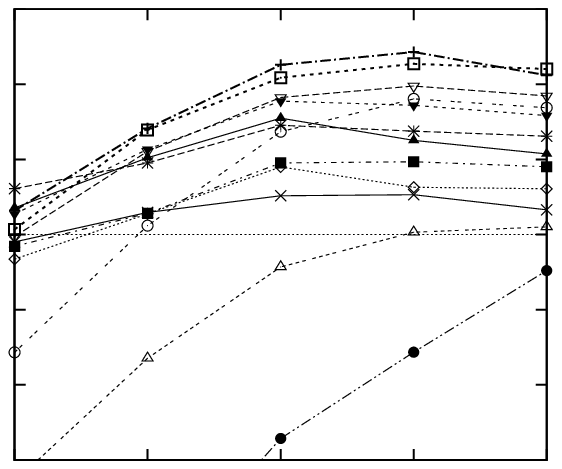} & \input{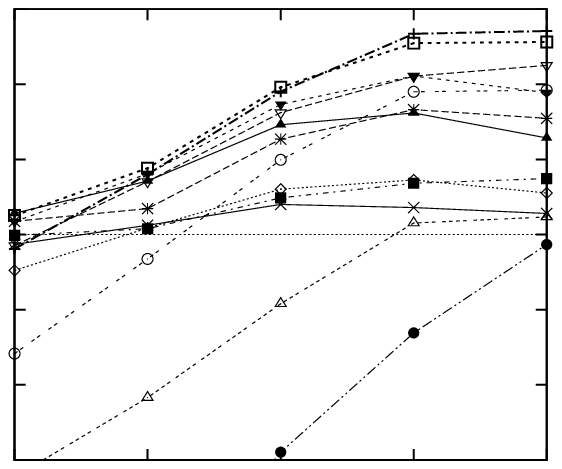}\\
MLP & Random Forest\\
\input{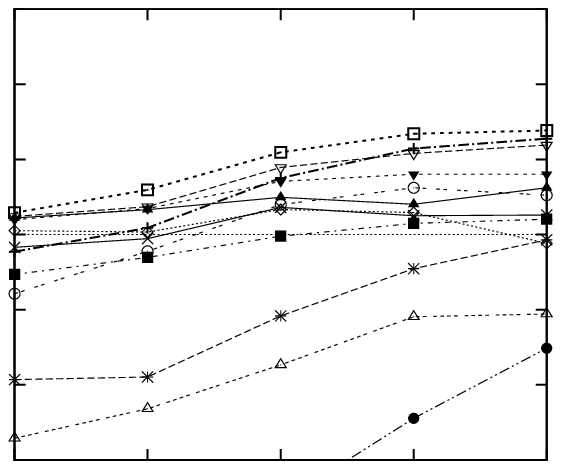} & \input{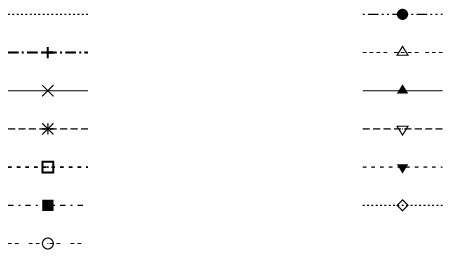}\\
RIPPER\\
\end{tabular}
\caption{Graphs for each learning algorithm comparing the average percent reduction in error achieved when using the investigated noise handling techniques.}
\label{figure:artificialNoise}
\end{figure}

Despite this, most previous work in noise handling has shown significant improvement when there is a high amount of noise in the data.
As there is no way to determine if an instance is noisy or mislabeled without the use of a domain expert, most previous work has added artificial noise to show the impact of noise and how handling noise improves the accuracy.
Generally, once there are large amounts of noise, using a noise handling approach significantly increases the classification accuracy.
In addition to not adding artificial noise, we also examine the effectiveness of the noise handling techniques with the addition of artificial noise.
Figure \ref{figure:artificialNoise} graphs the percent reduction in error for each learning algorithm and noise handling method for the considered artificial noise levels of 0\%, 10\%, 20\%, 30\%, and 40\%.
The percent reduction in error (\%RE) is the percentage of error that is reduced when a noise handling technique is used compared to the error obtained with no noise handling:
\begin{equation}
\%RE = \frac{noise-orig}{100-orig} \nonumber
\end{equation}
where $noise$ is the accuracy achieved when using a noise handling technique and $orig$ is the accuracy obtained when not using a noise handling technique.
The x-axis represents no change in error from not using a noise handling technique, negative values represent an increase in error when using a noise handling technique and positive values represent a decrease in error compared to not using noise handling.
In general, the reduction in the percentage of error increases as the amount of noise increases.
With the exception of ANR and the classification filter all of the noise handling techniques significantly increase the classification accuracy as the noise level increases.
In the cases of ANR and the classification filter, a broad application of the techniques is not appropriate although they may be beneficial in specific tasks.
The decrease in error is greatest for $\mathcal{L}$-weighting and $\mathcal{L}$-filtering for each learning algorithm.
Recall that the learning algorithms in $\mathcal{L}$ were chosen to be diverse so as to represent more of the hypothesis space $\mathcal{H}$.
This suggests that a better estimation of $p(y|x)$ produces more significant results for instance weighting and filtering.
This is shown empirically as $\mathcal{L}$-weighting and $\mathcal{L}$-filtering have the greatest reduction in error for each learning algorithm (Figure \ref{figure:artificialNoise}).
However, there is an obvious trade-off since obtaining a more accurate estimate of $p(y|x)$ requires more computational resources.
The accuracies and \textit{p}-values for each learning algorithm and noise handling technique are shown in Table \ref{table:LAsig} in \ref{section:sigTables}.

Examining the performance of the considered learning algorithms, we note that MLPs and random forests generally achieve the highest classification accuracy and may be the most tolerant to the noise inherent in each data set.
However, MLPs and random forests also appear to be the least robust as they obtain the lowest average classification accuracy when more than 10\% of the instances are corrupted with noise.
With no artificial noise, MLPs and random forests achieve about 81\% accuracy.
With 20\% artificial noise, the average accuracy decreases to about 72\%.
On the other hand, C4.5, IB5, and RIPPER achieve an average accuracy of about 79\% with no artificial noise and an average accuracy of about 74\% with 20\% artificial noise.
With high degrees of noise, the built-in noise handling mechanisms of learning algorithms become more beneficial.

\subsection{Comparison of Noise Handling Techniques}
We next compare the NICD noise handling techniques with the techniques that do not use classifier diversity.
We first compare different instance weighting schemes.
Table \ref{table:weightingSchemes} compares $\mathcal{L}$-weighting using the learning algorithms listed in Table \ref{table:LA} with biased weighting (B-W), and PWEM.
The values in bold represent the cases where $\mathcal{L}$-weighting achieves significantly higher classification accuracy than the compared weighting method.
$\mathcal{L}$-weighting significantly outperforms the other weighting schemes in most cases: 24 out of the 25 cases for PWEM and 20 out of the 25 for B-W. 
Biased-weighting and PWEM never achieve a significantly higher classification accuracy than $\mathcal{L}$-weighting.

\begin{table}
\centering
\caption{A comparison of the effect of the investigated instance weighting methods on the considered learning algorithm averaged over the 54 data sets. 
Bold \textit{p}-values represent cases where $\mathcal{L}$-weighting achieves significantly higher accuracy than the compared technique.}
\setlength{\tabcolsep}{0.6pt}
\small
\begin{tabular}{l|ccccc||ccccc}
\multicolumn{1}{c}{} & \multicolumn{5}{c}{C4dot5} & \multicolumn{5}{c}{IB5} \\
noise & 0 & 0.1 & 0.2 & 0.3 & 0.4 &  0 & 0.1 & 0.2 & 0.3 & 0.4 \\
\hline
$\mathcal{L}$-W & 78.36 & 77.39 & 76.44 & 74.32 & 70.84 & 78.72 & 78.09 & 76.77 & 74.8 & 70.30 \\
\hline
B-W & 79.29 & 77.22 & 75.28 & 71.06 & 65.74 & 78.34 & 77.41 & 75.39 & 71.59 & 64.92 \\
\textit{p}-val & 0.599 & 0.101 & \textbf{0.002} & $\mathbf{0.000}$ & $\mathbf{0.000}$ & \textbf{0.002} & $\mathbf{0.000}$ & $\mathbf{0.000}$ & $\mathbf{0.000}$ & $\mathbf{0.000}$ \\
Count & 26,4,24 & 33,2,19 & 38,0,16 & 45,0,9 & 43,0,10 & 34,7,13 & 41,3,10 & 43,1,10 & 46,1,7 & 44,1,8 \\
\hline
PWEM & 76.41 & 74.50 & 73.34 & 70.94 & 68.28 & 78.02 & 77.54 & 76.12 & 73.56 & 67.18 \\
\textit{p}-val & \textbf{0.003} & $\mathbf{0.000}$ & $\mathbf{0.000}$ & $\mathbf{0.000}$ & \textbf{0.001} & \textbf{0.007} & \textbf{0.005} & $\mathbf{0.000}$ & $\mathbf{0.000}$ & $\mathbf{0.000}$ \\
Count & 35,4,15 & 40,3,11 & 39,4,11 & 41,0,13 & 37,0,16 & 34,4,16 & 36,2,16 & 37,1,16 & 37,3,14 & 36,3,14 \\
 \multicolumn{1}{c}{} & \multicolumn{5}{c}{MLP} & \multicolumn{5}{c}{Random Forest} \\
noise & 0 & 0.1 & 0.2 & 0.3 & 0.4 &  0 & 0.1 & 0.2 & 0.3 & 0.4 \\
\hline
$\mathcal{L}$-W & 82.26 & 80.71 & 78.64 & 75.33 & 69.12 & 80.82 & 79.73 & 78.08 & 75.87 & 70.7 \\
\hline
B-W & 81.49 & 78.19 & 73.84 & 69.14 & 62.1 & 80.94 & 78.24 & 74.01 & 68.25 & 60.93 \\
\textit{p}-val & \textbf{0.042} & $\mathbf{0.000}$ & $\mathbf{0.000}$ & $\mathbf{0.000}$ & $\mathbf{0.000}$ & 0.523 & $\mathbf{0.000}$ & $\mathbf{0.000}$ & $\mathbf{0.000}$ & $\mathbf{0.000}$ \\
Count & 34,2,18 & 42,1,11 & 47,1,6 & 48,0,6 & 47,1,5 & 25,1,28 & 42,2,10 & 48,0,6 & 48,1,4 & 48,1,4 \\
\hline
PWEM & 82.79 & 79.67 & 76.42 & 71.9 & 65.95 & 81.51 & 78.73 & 76.37 & 72.54 & 66.03 \\
\textit{p}-val & \textbf{0.019} & $\mathbf{0.000}$ & $\mathbf{0.000}$ & $\mathbf{0.000}$ & $\mathbf{0.000}$ & 0.207 & \textbf{0.003} & $\mathbf{0.000}$ & $\mathbf{0.000}$ & $\mathbf{0.000}$ \\
Count & 34,4,16 & 38,2,14 & 39,1,14 & 42,0,12 & 41,0,12 & 33,4,17 & 34,4,16 & 40,4,10 & 47,1,5 & 48,1,4 \\
 \multicolumn{1}{c}{} & \multicolumn{5}{c}{RIPPER} \\
noise & 0 & 0.1 & 0.2 & 0.3 & 0.4 &\\
\cline{1-6}
$\mathcal{L}$-W & 77.86 & 76.62 & 75.53 & 73.38 & 69.53 \\
\cline{1-6}
B-W & 77.98 & 76.28 & 74.5 & 70.7 & 65.97 \\
\textit{p}-val & 0.402 & 0.085 & \textbf{0.002} & $\mathbf{0.000}$ & $\mathbf{0.000}$ \\
Count & 27,3,24 & 31,2,21 & 36,3,15 & 45,2,6 & 44,0,9 \\
\cline{1-6}
PWEM & 74.17 & 71.94 & 70.68 & 68.57 & 64.82 \\
\textit{p}-val & $\mathbf{0.000}$ & $\mathbf{0.000}$ & $\mathbf{0.000}$ & $\mathbf{0.000}$ & $\mathbf{0.000}$ \\
Count & 36,4,14 & 45,2,7 & 45,4,5 & 39,2,12 & 40,1,12 \\  
\end{tabular}
\label{table:weightingSchemes}
\end{table}

\begin{table}
\centering
\caption{A comparison of $\mathcal{L}$-filtering with the other filtering techniques for the 5 considered learning algorithms averaged over the 54 data sets.
Bold \textit{p}-values represent cases where $\mathcal{L}$-filtering achieves significantly higher accuracy than the compared technique.}
\setlength{\tabcolsep}{2.25pt}
\small
\begin{tabular}{l|ccccc||ccccc}
 \multicolumn{1}{c}{} & \multicolumn{5}{c}{C4dot5} & \multicolumn{5}{c}{IB5}\\ 
noise & 0 & 0.1 & 0.2 & 0.3 & 0.4 & 0 & 0.1 & 0.2 & 0.3 & 0.4 \\
\hline
$\mathcal{L}$-F & 79.55 & 78.35 & 76.65 & 73.42 & 69.14 & 79.4 & 78.35 & 76.62 & 74.38 & 69.67 \\
\hline
B-F & 79.34 & 77.49 & 75.17 & 71.32 & 65.58 & 76.99 & 75.98 & 74.33 & 71.82 & 66.12 \\
\textit{p}-val & \textbf{0.016} & $\textbf{0.000}$ & $\textbf{0.000}$ & $\textbf{0.000}$ & $\textbf{0.000}$ & $\textbf{0.000}$ & $\textbf{0.000}$ & $\textbf{0.000}$ & $\textbf{0.000}$ & $\textbf{0.000}$ \\
Count & \begin{footnotesize}31,8,15\end{footnotesize} & \begin{footnotesize}42,1,11\end{footnotesize} & \begin{footnotesize}44,4,6\end{footnotesize} & \begin{footnotesize}44,1,9\end{footnotesize} & \begin{footnotesize}46,0,8\end{footnotesize} & \begin{footnotesize}41,4,9\end{footnotesize} & \begin{footnotesize}45,0,9\end{footnotesize} & \begin{footnotesize}43,0,11\end{footnotesize} & \begin{footnotesize}37,1,16\end{footnotesize} & \begin{footnotesize}39,1,14\end{footnotesize} \\
\hline
CVC & 79.41 & 77.35 & 74.6 & 70.11 & 63.64 & 79.37 & 78.38 & 76.5 & 73.21 & 66.36 \\
\textit{p}-val & 0.126 & $\textbf{0.000}$ & $\textbf{0.000}$ & $\textbf{0.000}$ & $\textbf{0.000}$ & 0.175 & 0.211 & \textbf{0.028} & $\textbf{0.000}$ & $\textbf{0.000}$ \\
Count & \begin{footnotesize}25,8,21\end{footnotesize} & \begin{footnotesize}43,1,10\end{footnotesize} & \begin{footnotesize}50,1,3\end{footnotesize} & \begin{footnotesize}47,3,4\end{footnotesize} & \begin{footnotesize}48,2,4\end{footnotesize} & \begin{footnotesize}26,7,21\end{footnotesize} & \begin{footnotesize}26,5,23\end{footnotesize} & \begin{footnotesize}34,2,18\end{footnotesize} & \begin{footnotesize}37,2,15\end{footnotesize} & \begin{footnotesize}39,2,13\end{footnotesize} \\
\hline
Ens & 79.53 & 77.88 & 76.15 & 73.62 & 68.66 & 79.06 & 77.99 & 76.33 & 73.51 & 68.92 \\
\textit{p}-val & 0.438 & \textbf{0.033} & \textbf{0.003} & 0.544 & 0.124 & \textbf{0.017} & 0.075 & 0.061 & \textbf{0.007} & 0.052 \\
Count & \begin{footnotesize}27,2,25\end{footnotesize} & \begin{footnotesize}34,3,17\end{footnotesize} & \begin{footnotesize}35,3,16\end{footnotesize} & \begin{footnotesize}26,1,27\end{footnotesize} & \begin{footnotesize}30,1,23\end{footnotesize} & \begin{footnotesize}32,7,15\end{footnotesize} & \begin{footnotesize}29,3,22\end{footnotesize} & \begin{footnotesize}29,4,21\end{footnotesize} & \begin{footnotesize}34,1,19\end{footnotesize} & \begin{footnotesize}32,0,22\end{footnotesize} \\
\multicolumn{1}{c}{} & \multicolumn{5}{c}{MLP} & \multicolumn{5}{c}{RandForest}\\ 
noise & 0 & 0.1 & 0.2 & 0.3 & 0.4 & 0 & 0.1 & 0.2 & 0.3 & 0.4 \\
\hline
$\mathcal{L}$-F & 81.80 & 80.66 & 78.17 & 74.82 & 69.44 & 81.66 & 79.91 & 78.23 & 75.45 & 70.11 \\
\hline
B-F & 81.39 & 78.16 & 75.04 & 70.57 & 64.35 & 81.16 & 78.13 & 74.25 & 69.32 & 62.8 \\
\textit{p}-val & 0.210 & $\textbf{0.000}$ & $\textbf{0.000}$ & $\textbf{0.000}$ & $\textbf{0.000}$ & 0.096 & $\textbf{0.000}$ & $\textbf{0.000}$ & $\textbf{0.000}$ & $\textbf{0.000}$ \\
Count & \begin{footnotesize}25,4,25\end{footnotesize} & \begin{footnotesize}46,0,8\end{footnotesize} & \begin{footnotesize}44,1,9\end{footnotesize} & \begin{footnotesize}44,0,10\end{footnotesize} & \begin{footnotesize}46,1,7\end{footnotesize} & \begin{footnotesize}28,1,25\end{footnotesize} & \begin{footnotesize}42,1,11\end{footnotesize} & \begin{footnotesize}47,1,6\end{footnotesize} & \begin{footnotesize}50,0,4\end{footnotesize} & \begin{footnotesize}50,0,4\end{footnotesize} \\
\hline
CVC & 82.33 & 79.83 & 76.68 & 71.5 & 65.02 & 81.72 & 79.53 & 76.88 & 72.4 & 64.97 \\
\textit{p}-val & 0.905 & \textbf{0.002} & $\textbf{0.000}$ & $\textbf{0.000}$ & $\textbf{0.000}$ & 0.786 & 0.070 & $\textbf{0.000}$ & $\textbf{0.000}$ & $\textbf{0.000}$ \\
Count & \begin{footnotesize}22,3,29\end{footnotesize} & \begin{footnotesize}33,2,19\end{footnotesize} & \begin{footnotesize}43,2,9\end{footnotesize} & \begin{footnotesize}46,0,8\end{footnotesize} & \begin{footnotesize}48,1,5\end{footnotesize} & \begin{footnotesize}23,4,27\end{footnotesize} & \begin{footnotesize}31,2,21\end{footnotesize} & \begin{footnotesize}40,1,13\end{footnotesize} & \begin{footnotesize}47,1,6\end{footnotesize} & \begin{footnotesize}47,1,6\end{footnotesize} \\
\hline
Ens & 81.62 & 80.02 & 77.44 & 73.85 & 68.04 & 80.91 & 79.52 & 77.32 & 74.01 & 68.86 \\
\textit{p}-val & \textbf{0.009} & $\textbf{0.000}$ & \textbf{0.009} & \textbf{0.020} & \textbf{0.006} & $\textbf{0.000}$ & 0.069 & $\textbf{0.000}$ & $\textbf{0.000}$ & \textbf{0.006} \\
Count & \begin{footnotesize}32,3,19\end{footnotesize} & \begin{footnotesize}36,2,16\end{footnotesize} & \begin{footnotesize}32,2,20\end{footnotesize} & \begin{footnotesize}30,1,23\end{footnotesize} & \begin{footnotesize}34,0,20\end{footnotesize} & \begin{footnotesize}38,1,15\end{footnotesize} & \begin{footnotesize}31,1,22\end{footnotesize} & \begin{footnotesize}39,2,13\end{footnotesize} & \begin{footnotesize}38,1,15\end{footnotesize} & \begin{footnotesize}34,0,20\end{footnotesize} \\
\multicolumn{1}{c}{} & \multicolumn{5}{c}{RIPPER}\\   
noise & 0 & 0.1 & 0.2 & 0.3 & 0.4 \\  
\cline{1-6}
$\mathcal{L}$-F & 78.98 & 77.82 & 76.43 & 73.97 & 69.9 \\  
\cline{1-6}
B-F & 77.20 & 75.70 & 73.48 & 70.39 & 65.78 \\  
\textit{p}-val & $\textbf{0.000}$ & $\textbf{0.000}$ & $\textbf{0.000}$ & $\textbf{0.000}$ & $\textbf{0.000}$ \\  
Count & \begin{footnotesize}37,3,14\end{footnotesize} & \begin{footnotesize}44,1,9\end{footnotesize} & \begin{footnotesize}46,0,8\end{footnotesize} & \begin{footnotesize}44,1,9\end{footnotesize} & \begin{footnotesize}44,2,8\end{footnotesize} \\  
\cline{1-6}
CVC & 78.83 & 77.19 & 74.85 & 71.15 & 67.24 \\  
\textit{p}-val & 0.257 & \textbf{0.013} & $\textbf{0.000}$ & $\textbf{0.000}$ & $\textbf{0.000}$ \\  
Count & \begin{footnotesize}26,4,24\end{footnotesize} & \begin{footnotesize}32,1,21\end{footnotesize} & \begin{footnotesize}40,2,12\end{footnotesize} & \begin{footnotesize}50,1,3\end{footnotesize} & \begin{footnotesize}41,1,12\end{footnotesize} \\  
\cline{1-6}
Ens & 78.87 & 77.29 & 75.9 & 73.19 & 69.22 \\  
\textit{p}-val & 0.201 & 0.072 & \textbf{0.003} & \textbf{0.031} & 0.323 \\  
Count & \begin{footnotesize}28,3,23\end{footnotesize} & \begin{footnotesize}32,4,18\end{footnotesize} & \begin{footnotesize}36,1,17\end{footnotesize} & \begin{footnotesize}32,1,21\end{footnotesize} & \begin{footnotesize}24,1,29\end{footnotesize} \\  

\end{tabular}
\label{table:filtering}
\end{table}

The results are similar for filtering.
Table \ref{table:filtering} compares $\mathcal{L}$-filtering ($\mathcal{L}$-F) with 1) the biased filter (B-F), 2) the cross-validated committees filter (CVC), and 3) the ensemble filter (Ens).
The CVC and Ens filters were chosen as representative filtering algorithms as they achieved the highest and most significant increase in classification accuracy.
For filtering, no filtering approach significantly outperforms the $\mathcal{L}$-filter, yet the $\mathcal{L}$-filter achieves significantly higher classification accuracy in most cases (bold \textit{p}-values in Table \ref{table:filtering}).

\begin{table}[t]
\centering
\caption{A comparison of the $\mathcal{L}$-ensemble with the 3-Ensemble and with Weighted $\mathcal{L}$-Ensemble and Filtered $\mathcal{L}$-Ensemble averaged over the 54 data sets.
Bold \textit{p}-values represent cases where the $\mathcal{L}$-ensemble achieves significantly higher accuracy; underlined, gray cells represent cases where the $\mathcal{L}$-ensemble achieves significantly lower accuracy.}
\setlength{\tabcolsep}{4.5pt}
\begin{tabular}{l|ccccc}
Noise & 0\% & 0.1\% & 0.2\% & 0.3\% & 0.4\% \\
\hline
$\mathcal{L}$-Ensemble & 83.36 & 82.06 & 79.87 & 77.09 & 72.04 \\
\hline
3-Ensemble & 82.64 & 80.75 & 77.14 & 72.37 & 64.94 \\
\textit{p}-val & \textbf{0.001} & $\mathbf{<0.001}$ & $\mathbf{<0.001}$ & $\mathbf{<0.001}$ & $\mathbf{<0.001}$ \\
Count & 36,3,15 & 42,1,11 & 46,1,7 & 49,0,5 & 50,0,4 \\
\hline
Weighted $\mathcal{L}$-Ensemble & 82.03 & 81.92 & 79.62 & 77.58 & 73.81 \\
\textit{p}-val & $\mathbf{<0.001}$ & 0.165 & 0.550 & \cellcolor{gray}\underline{0.963} & \cellcolor{gray}\underline{1.000} \\
Count & 41,2,11 & 27,3,24 & 24,0,30 & 19,1,34 & 16,1,37 \\
\hline
Filtered $\mathcal{L}$-Ensemble  & 82.98 & 81.66 & 79.87 & 77.74 & 73.29 \\
\textit{p}-val & \textbf{0.003} & \textbf{0.013} & 0.842 & \cellcolor{gray}\underline{1.000} & \cellcolor{gray}\underline{1.000} \\
Count & 33,6,15 & 30,4,20 & 22,3,29 & 11,0,43 & 12,0,42 \\
\end{tabular}
\label{table:ensembleComp}
\end{table}

Table \ref{table:ensembleComp} compares the $\mathcal{L}$-ensemble with a voting ensemble composed of a subset of three learning algorithms from $\mathcal{L}$ as the base classifiers (3-Ensem-ble). 
The 3-Ensemble is composed of the 5-NN, MLP, and random forest learning algorithms.
The three learning algorithms in the 3-Ensemble were chosen 
because they obtain high classification accuracy and are diverse (see Section \ref{section:estimatingP}).
In other words, we selected the learning algorithm from three clusters that had the highest classification accuracy.
We chose not to use the cluster that contains the LWL and RIPPER learning algorithms as those typically achieved the lowest classification accuracy of the nine investigated learning algorithms.
Additionally, the classifier scores for random forests have the highest correlation with $p(y|x)$ \cite{Smith2012_IH}.
Using only three learning algorithms gives a limited view of the hypothesis space since not as many learning algorithms are used. 
The $\mathcal{L}$-ensemble significantly outperforms the 3-ensemble at all noise levels.
Thus, a better sampling of $\mathcal{H}$ is beneficial in this case. 

Table \ref{table:ensembleComp} also compares the $\mathcal{L}$-ensemble with a weighted $\mathcal{L}$-ensemble and a filtered $\mathcal{L}$-ensemble.
The base classifiers for the weighted $\mathcal{L}$-ensemble are the biased-weighted classifiers used earlier.
Likewise, the base classifiers for the filtered $\mathcal{L}$-ensemble are the biased-filtered classifiers.
With no artificial noise added to the data set, the $\mathcal{L}$-ensemble without noise handling achieves significantly higher classification accuracy than using noise handling in the base classifiers.
This is not surprising as noise handling often decreases the classification accuracy when there is little to no noise in a data set (see Table \ref{table:noNoise}).
However, with higher noise levels, the weighted and filtered $\mathcal{L}$-ensembles significantly increase the classification accuracy.
Up to about 20\% to 30\% class noise the $\mathcal{L}$-ensemble is robust to class noise.

\begin{table}
\centering
\caption{A comparison of the $\mathcal{L}$-ensemble against the investigated noise handling approaches for the 5 considered learning algorithm averaged over the 54 data sets.}
\setlength{\tabcolsep}{1.4pt}
\begin{tabular}{l|ccccc||ccccc}
noise  & 0\% & 10\% & 20\% & 30\% & 40\% & 0\% & 10\% & 20\% & 30\% & 40\% \\
\hline
$\mathcal{L}$-Ens & 83.36 & 82.06 & 79.87 & 77.09 & 72.04 & \\
\hline
\multicolumn{11}{c}{}\\
& \multicolumn{5}{c||}{C4.5}& \multicolumn{5}{c}{IB5} \\
\hline
$\mathcal{L}$-Weight & 78.36 & 77.39 & 76.44 & 74.32 & 70.84 & 78.72 & 78.09 & 76.77 & 74.8 & 70.3 \\
\textit{p}-val & $\textbf{0.000}$ & $\textbf{0.000}$ & $\textbf{0.000}$ & $\textbf{0.000}$ & 0.168 & $\textbf{0.000}$ & $\textbf{0.000}$ & $\textbf{0.000}$ & $\textbf{0.000}$ & \textbf{0.007} \\
Count & \begin{footnotesize}46,1,7\end{footnotesize} & \begin{footnotesize}44,0,10\end{footnotesize} & \begin{footnotesize}39,0,15\end{footnotesize} & \begin{footnotesize}37,0,17\end{footnotesize} & \begin{footnotesize}29,0,24\end{footnotesize} & \begin{footnotesize}44,4,6\end{footnotesize} & \begin{footnotesize}44,0,10\end{footnotesize} & \begin{footnotesize}38,2,14\end{footnotesize} & \begin{footnotesize}37,1,16\end{footnotesize} & \begin{footnotesize}33,3,17\end{footnotesize} \\
\hline
$\mathcal{L}$-Filter & 79.55 & 78.35 & 76.65 & 73.42 & 69.14 & 79.4 & 78.35 & 76.62 & 74.38 & 69.67 \\
\textit{p}-val & $\textbf{0.000}$ & $\textbf{0.000}$ & $\textbf{0.000}$ & $\textbf{0.000}$ & $\textbf{0.000}$ & $\textbf{0.000}$ & $\textbf{0.000}$ & $\textbf{0.000}$ & $\textbf{0.000}$ & $\textbf{0.000}$ \\
Count & \begin{footnotesize}45,3,6\end{footnotesize} & \begin{footnotesize}44,0,10\end{footnotesize} & \begin{footnotesize}42,0,12\end{footnotesize} & \begin{footnotesize}46,1,7\end{footnotesize} & \begin{footnotesize}37,0,17\end{footnotesize} & \begin{footnotesize}47,2,5\end{footnotesize} & \begin{footnotesize}42,2,10\end{footnotesize} & \begin{footnotesize}42,2,10\end{footnotesize} & \begin{footnotesize}42,1,11\end{footnotesize} & \begin{footnotesize}36,0,18\end{footnotesize} \\
\hline
Ens Filt & 79.53 & 77.88 & 76.15 & 73.62 & 68.66 & 79.06 & 77.99 & 76.33 & 73.51 & 68.92 \\
\textit{p}-val & $\textbf{0.000}$ & $\textbf{0.000}$ & $\textbf{0.000}$ & $\textbf{0.000}$ & $\textbf{0.000}$ & $\textbf{0.000}$ & $\textbf{0.000}$ & $\textbf{0.000}$ & $\textbf{0.000}$ & $\textbf{0.000}$ \\
Count & \begin{footnotesize}46,1,7\end{footnotesize} & \begin{footnotesize}45,0,9\end{footnotesize} & \begin{footnotesize}42,0,12\end{footnotesize} & \begin{footnotesize}43,0,11\end{footnotesize} & \begin{footnotesize}40,0,14\end{footnotesize} & \begin{footnotesize}44,1,9\end{footnotesize} & \begin{footnotesize}45,0,9\end{footnotesize} & \begin{footnotesize}42,0,12\end{footnotesize} & \begin{footnotesize}41,1,12\end{footnotesize} & \begin{footnotesize}37,0,17\end{footnotesize} \\
\hline
\multicolumn{11}{c}{}\\
& \multicolumn{5}{c||}{MLP}& \multicolumn{5}{c}{Random Forest} \\
\hline
$\mathcal{L}$-Weight & 82.26 & 80.71 & 78.64 & 75.33 & 69.12 & 80.82 & 79.73 & 78.08 & 75.87 & 70.7 \\
\textit{p}-val & $\textbf{0.000}$ & $\textbf{0.000}$ & \textbf{0.001} & $\textbf{0.000}$ & $\textbf{0.000}$ & $\textbf{0.000}$ & $\textbf{0.000}$ & $\textbf{0.000}$ & \textbf{0.002} & \textbf{0.006} \\
Count & \begin{footnotesize}37,3,14\end{footnotesize} & \begin{footnotesize}42,0,12\end{footnotesize} & \begin{footnotesize}36,2,16\end{footnotesize} & \begin{footnotesize}36,1,17\end{footnotesize} & \begin{footnotesize}36,0,17\end{footnotesize} & \begin{footnotesize}44,0,10\end{footnotesize} & \begin{footnotesize}41,3,10\end{footnotesize} & \begin{footnotesize}41,2,11\end{footnotesize} & \begin{footnotesize}35,1,17\end{footnotesize} & \begin{footnotesize}30,0,23\end{footnotesize} \\
\hline
$\mathcal{L}$-Filter & 81.8 & 80.66 & 78.17 & 74.82 & 69.44 & 81.66 & 79.91 & 78.23 & 75.45 & 70.11 \\
\textit{p}-val & $\textbf{0.000}$ & $\textbf{0.000}$ & $\textbf{0.000}$ & $\textbf{0.000}$ & $\textbf{0.000}$ & $\textbf{0.000}$ & $\textbf{0.000}$ & $\textbf{0.000}$ & $\textbf{0.000}$ & $\textbf{0.000}$ \\
Count & \begin{footnotesize}40,5,9\end{footnotesize} & \begin{footnotesize}39,1,14\end{footnotesize} & \begin{footnotesize}41,0,13\end{footnotesize} & \begin{footnotesize}40,2,12\end{footnotesize} & \begin{footnotesize}36,0,18\end{footnotesize} & \begin{footnotesize}41,4,9\end{footnotesize} & \begin{footnotesize}49,0,5\end{footnotesize} & \begin{footnotesize}45,0,9\end{footnotesize} & \begin{footnotesize}42,1,11\end{footnotesize} & \begin{footnotesize}40,0,14\end{footnotesize} \\
\hline
Ens Filt & 81.62 & 80.02 & 77.44 & 73.85 & 68.04 & 80.91 & 79.52 & 77.32 & 74.01 & 68.86 \\
\textit{p}-val & $\textbf{0.000}$ & $\textbf{0.000}$ & $\textbf{0.000}$ & $\textbf{0.000}$ & $\textbf{0.000}$ & $\textbf{0.000}$ & $\textbf{0.000}$ & $\textbf{0.000}$ & $\textbf{0.000}$ & $\textbf{0.000}$ \\
Count & \begin{footnotesize}42,1,11\end{footnotesize} & \begin{footnotesize}44,2,8\end{footnotesize} & \begin{footnotesize}46,0,8\end{footnotesize} & \begin{footnotesize}46,2,6\end{footnotesize} & \begin{footnotesize}46,1,7\end{footnotesize} & \begin{footnotesize}45,1,8\end{footnotesize} & \begin{footnotesize}47,0,7\end{footnotesize} & \begin{footnotesize}47,1,6\end{footnotesize} & \begin{footnotesize}44,4,6\end{footnotesize} & \begin{footnotesize}45,2,7\end{footnotesize} \\
\multicolumn{11}{c}{}\\
& \multicolumn{5}{c||}{RIPPER}& \\
\cline{1-6}
$\mathcal{L}$-Weight & 77.86 & 76.62 & 75.53 & 73.38 & 69.53 \\
\textit{p}-val & $\textbf{0.000}$ & $\textbf{0.000}$ & $\textbf{0.000}$ & $\textbf{0.000}$ & \textbf{0.017} \\
Count & \begin{footnotesize}47,2,5\end{footnotesize} & \begin{footnotesize}48,0,6\end{footnotesize} & \begin{footnotesize}41,1,12\end{footnotesize} & \begin{footnotesize}39,0,14\end{footnotesize} & \begin{footnotesize}32,1,20\end{footnotesize} \\
\cline{1-6}
$\mathcal{L}$-Filter & 78.98 & 77.82 & 76.43 & 73.97 & 69.9 \\
\textit{p}-val & $\textbf{0.000}$ & $\textbf{0.000}$ & $\textbf{0.000}$ & $\textbf{0.000}$ & $\textbf{0.000}$ \\
Count & \begin{footnotesize}47,2,5\end{footnotesize} & \begin{footnotesize}49,0,5\end{footnotesize} & \begin{footnotesize}44,0,10\end{footnotesize} & \begin{footnotesize}42,1,11\end{footnotesize} & \begin{footnotesize}37,1,16\end{footnotesize} \\
\cline{1-6}
Ens Filt& 78.87 & 77.29 & 75.9 & 73.19 & 69.22 \\
\textit{p}-val & $\textbf{0.000}$ & $\textbf{0.000}$ & $\textbf{0.000}$ & $\textbf{0.000}$ & \textbf{0.002} \\
Count & \begin{footnotesize}50,1,3\end{footnotesize} & \begin{footnotesize}48,0,6\end{footnotesize} & \begin{footnotesize}45,1,8\end{footnotesize} & \begin{footnotesize}40,1,13\end{footnotesize} & \begin{footnotesize}35,1,18\end{footnotesize} \\

\end{tabular}
\label{table:voting}
\end{table}

For completeness, Table \ref{table:voting} compares the $\mathcal{L}$-ensemble with $\mathcal{L}$-weighting, $\mathcal{L}$-filtering, and the ensemble filter.
These noise handling methods were chosen to be representative as they achieved the highest and most significant gains in classification accuracy in a single learning algorithm.
Despite the significant increase in classification accuracy over the original learning algorithm by the noise handling method, the $\mathcal{L}$-ensemble achieves significantly higher classification accuracy over all of the considered learning algorithms at all of the noise levels.
This again demonstrates the $\mathcal{L}$-ensemble's robustness to label noise.

The increase in accuracy by the voting ensemble is not too surprising as ensembles have been shown to perform better than any one of their constituent base classifiers \cite{Dietterich2000}.
We focus our attention on the MLP and random forest learning algorithms since they obtain the highest classification accuracy of the considered learning algorithms.
Although the increase in accuracy by a voting ensemble is significant, it is generally within 1 to 2 percent of the average accuracy achieved by a MLP or a random forest with $\mathcal{L}$-weighting or $\mathcal{L}$-filtering.
Noise handling does increase the classification accuracy, but the $\mathcal{L}$-ensemble is a safe choice to use in general.
However, when compared to the 3-ensemble (in which MLP is one of the base classifiers) with no artificial noise, $\mathcal{L}$-weighting in a MLP is statistically equivalent.
The MLP achieves an average accuracy of 82.26\% compared to 82.45\% by the 3-ensemble.
With higher levels of noise, the $\mathcal{L}$-weighted MLP achieves significantly higher average accuracy.
Thus, representing $\mathcal{H}$ whether by weighting or by ensembling a set of base classifiers has important ramifications.
Detailed results comparing $\mathcal{L}$-weighting and $\mathcal{L}$-filtering with the 3-ensemble are provided in Table \ref{table:voting_3} in \ref{appendix_3-ens}.
Further investigation of the diversity of the base learning algorithms and its impact on noise and classification accuracy is being pursued as a topic of on-going research.

\section{Conclusions}
\label{section:conclusions}

In this paper we examined handling class noise using the hypotheses from a diverse set of learning algorithms.
We introduced \textit{noise identification using classifier diversity} (NICD) which uses a diverse set of learning algorithms to approximate $p(y|x)$.
Using this diverse set of learning algorithms, we examined NICD as a filtering technique ($\mathcal{L}$-filtering), a weighting technique ($\mathcal{L}$-weighting), and as the base classifiers in a voting ensemble ($\mathcal{L}$-ensemble) on a set of 5 learning algorithms and 54 data sets.
We found that NICD significantly outperforms other less diverse noise handling techniques.
Our results suggest that a better estimate of $p(y|x)$ leads to better noise handling -- however, there is no way of actually knowing $p(y|x)$.
Compared to no noise handling, $\mathcal{L}$-weighting and $\mathcal{L}$-filtering often achieve a significantly higher classification accuracy and never achieve a significantly lower classification accuracy.
Thus, NICD is able to be applied across a broad set of data sets and learning algorithms.
They also significantly outperform the other considered noise handling techniques in most cases.
At worst, $\mathcal{L}$-weighting and $\mathcal{L}$-filtering are statistically equivalent to the other techniques.

Despite the increase in accuracy exhibited by $\mathcal{L}$-weighting and $\mathcal{L}$-filtering, the $\mathcal{L}$-ensemble achieves a significantly higher classification accuracy for all of the learning algorithms and noise handling techniques for all of the examined noise levels.
The $\mathcal{L}$-ensemble is a voting ensemble composed of a diverse set of learning algorithms (Section \ref{section:estimatingP}) with each learning algorithm having an equally-weighted vote.  
Thus, a voting ensemble exhibits a robustness to noise that the individual constituent learning algorithms do not possess.
The diversity of the base classifiers produces a robustness against label noise.



\bibliographystyle{elsarticle-num}

\begin{thebibliography}{10}
\expandafter\ifx\csname url\endcsname\relax
  \def\url#1{\texttt{#1}}\fi
\expandafter\ifx\csname urlprefix\endcsname\relax\def\urlprefix{URL }\fi
\expandafter\ifx\csname href\endcsname\relax
  \def\href#1#2{#2} \def\path#1{#1}\fi

\bibitem{Saez2013}
J.~A. S{\'a}ez, J.~Luengo, F.~Herrera, Predicting noise filtering efficacy with
  data complexity measures for nearest neighbor classification, Pattern
  Recognition 46~(1) (2013) 355--364.

\bibitem{Zhu2004}
X.~Zhu, X.~Wu, Class noise vs. attribute noise: a quantitative study of their
  impacts, Artificial Intelligence Review 22 (2004) 177--210.

\bibitem{Nettleton2010}
D.~F. Nettleton, A.~Orriols-Puig, A.~Fornells, A study of the effect of
  different types of noise on the precision of supervised learning techniques,
  Artificial Intelligence Review 33~(4) (2010) 275--306.

\bibitem{Frenay2014}
B.~Fr\'{e}nay, M.~Verleysen, Classification in the presence of label noise: a
  survey, IEEE Transactions on Neural Networks and Learning Systems (2014) in
  press, 25 pages.

\bibitem{Tomek1976}
I.~Tomek, An experiment with the edited nearest-neighbor rule, IEEE
  Transactions on Systems, Man, and Cybernetics 6 (1976) 448--452.

\bibitem{John95}
G.~H. John, Robust decision trees: Removing outliers from databases, in:
  Knowledge Discovery and Data Mining, 1995, pp. 174--179.

\bibitem{Smith_Eval}
M.~R. Smith, T.~Martinez, An extensive evaluation of filtering misclassified
  instances in supervised classification tasks, In submission.
\newline\urlprefix\url{http://arxiv.org/abs/1312.3970}

\bibitem{Dietterich2000}
T.~G. Dietterich, Ensemble methods in machine learning, in: Multiple Classifier
  Systems, Vol. 1857 of Lecture Notes in Computer Science, Springer, 2000, pp.
  1--15.

\bibitem{Brodley1999}
C.~E. Brodley, M.~A. Friedl, Identifying mislabeled training data, Journal of
  Artificial Intelligence Research 11 (1999) 131--167.

\bibitem{Verbaeten2003}
S.~Verbaeten, A.~{Van Assche}, Ensemble methods for noise elimination in
  classification problems, in: Proceedings of the 4th international conference
  on multiple classifier systems, 2003, pp. 317--325.

\bibitem{Khoshgoftaar2004_ensembleFilter}
T.~M. Khoshgoftaar, P.~Rebours, Generating multiple noise elimination filters
  with the ensemble-partitioning filter, in: Proceedings of the IEEE
  International Conference on Information Reuse and Integration, 2004, pp.
  369--375.

\bibitem{Hansen1990}
L.~K. Hansen, P.~Salamon, Neural network ensembles, IEEE Transactions on
  Pattern Analysis Machine Intelligence 12~(10) (1990) 993--1001.

\bibitem{Quinlan1993}
J.~R. Quinlan, C4.5: Programs for Machine Learning, Morgan Kaufmann, San Mateo,
  CA, USA, 1993.

\bibitem{bishop2006pattern}
C.~M. Bishop, N.~M. Nasrabadi, Pattern Recognition and Machine Learning,
  Vol.~1, springer New York, 2006.

\bibitem{Manwani2013}
N.~Manwani, P.~S. Sastry, Noise tolerance under risk minimization, IEEE
  Transactions on Cybernetics 43~(3) (2013) 1146--1151.

\bibitem{Schapire1990}
R.~E. Schapire, The strength of weak learnability, Machine Learning 5 (1990)
  197--227.

\bibitem{Freund1995}
Y.~Freund, Boosting a weak learning algorithm by majority, in: Proceedings of
  the Third Annual Workshop on Computational Learning Theory, 1990, pp.
  202--216.

\bibitem{Servedio2003}
R.~A. Servedio, Smooth boosting and learning with malicious noise, Journal of
  Machine Learning Research 4 (2003) 633--648.

\bibitem{Collobert2006}
R.~Collobert, F.~Sinz, J.~Weston, L.~Bottou, Trading convexity for scalability,
  in: Proceedings of the 23rd International Conference on Machine learning,
  2006, pp. 201--208.

\bibitem{Rosaloes2009}
R.~Rosales, G.~Fung, W.~Tong, Automatic discrimination of mislabeled training
  points for large margin classifiers, in: Snowbird Learning Workshop, 2009,
  pp. 1--2.

\bibitem{Lawrence2001}
N.~D. Lawrence, B.~Sch{\"o}lkopf, Estimating a kernel fisher discriminant in
  the presence of label noise, in: In Proceedings of the 18th International
  Conference on Machine Learning, 2001, pp. 306--313.

\bibitem{Gamberger2000}
D.~Gamberger, N.~Lavra\v{c}, S.~D\v{z}eroski, Noise detection and elimination
  in data preprocessing: Experiments in medical domains, Applied Artificial
  Intelligence 14~(2) (2000) 205--223.

\bibitem{Smith2011}
M.~R. Smith, T.~Martinez, Improving classification accuracy by identifying and
  removing instances that should be misclassified, in: Proceedings of the IEEE
  International Joint Conference on Neural Networks, 2011, pp. 2690--2697.

\bibitem{Wilson1972}
D.~L. Wilson, Asymptotic properties of nearest neighbor rules using edited
  data, IEEE Transactions on Systems, Man, and Cybernetics~(2-3) (1972)
  408--421.

\bibitem{Segata2009}
N.~Segata, E.~Blanzieri, P.~Cunningham, A scalable noise reduction technique
  for large case-based systems, in: Proceedings of the 8th International
  Conference on Case-Based Reasoning: Case-Based Reasoning Research and
  Development, 2009, pp. 328--342.

\bibitem{Xinchuan2003}
X.~Zeng, T.~R. Martinez, A noise filtering method using neural networks, in:
  Proc. of the int. Workshop of Soft Comput. Techniques in Instrumentation,
  Measurement and Related Applications, 2003.

\bibitem{Rebbapragada2007}
U.~Rebbapragada, C.~E. Brodley, Class noise mitigation through instance
  weighting, in: Proceedings of the 18th European Conference on Machine
  Learning, 2007, pp. 708--715.

\bibitem{zeng.ida2001}
X.~Zeng, T.~R. Martinez, An algorithm for correcting mislabeled data,
  Intelligent Data Analysis 5 (2001) 491--502.

\bibitem{Teng2000}
C.-M. Teng, Evaluating noise correction, in: PRICAI, 2000, pp. 188--198.

\bibitem{Teng2003}
C.~Teng, Combining noise correction with feature selection, in: Data
  Warehousing and Knowledge Discovery, Vol. 2737 of Lecture Notes in Computer
  Science, 2003, pp. 340--349.

\bibitem{Folleco2009}
A.~Folleco, T.~M. Khoshgoftaar, J.~V. Hulse, A.~Napolitano, Identifying
  learners robust to low quality data., Informatica 33~(3) (2009) 245--259.

\bibitem{Khoshgoftaar2010}
T.~M. Khoshgoftaar, J.~V. Hulse, A.~Napolitano, Supervised neural network
  modeling: an empirical investigation into learning from imbalanced data with
  labeling errors., IEEE Transactions on Neural Networks 21~(5) (2010)
  813--830.

\bibitem{Brown2005_JMLR}
G.~Brown, J.~L. Wyatt, P.~Tino, Managing diversity in regression ensembles.,
  Journal of Machine Learning Research 6 (2005) 1621--1650.

\bibitem{Brown2005}
G.~Brown, J.~L. Wyatt, R.~Harris, X.~Yao, Diversity creation methods: a survey
  and categorisation, Information Fusion 6~(1) (2005) 5--20.

\bibitem{Aksela2006}
M.~Aksela, J.~Laaksonen, Using diversity of errors for selecting members of a
  committee classifier, Pattern Recognition 39~(4) (2006) 608--623.

\bibitem{Shipp2002}
C.~A. Shipp, L.~Kuncheva, Relationships between combination methods and
  measures of diversity in combining classifiers., Information Fusion 3~(2)
  (2002) 135--148.

\bibitem{Kuncheva2003_ElusiveDiversity}
L.~Kuncheva, That elusive diversity in classifier ensembles, in: Pattern
  Recognition and Image Analysis, Vol. 2652 of Lecture Notes in Computer
  Science, 2003, pp. 1126--1138.

\bibitem{Ruta2005}
D.~Ruta, B.~Gabrys, Classifier selection for majority voting, Information
  Fusion 6~(1) (2005) 63--81.

\bibitem{Kuncheva2003}
L.~I. Kuncheva, C.~J. Whitaker, Measures of diversity in classifier ensembles
  and their relationship with the ensemble accuracy., Machine Learning 51~(2)
  (2003) 181--207.

\bibitem{Peterson2005}
A.~H. Peterson, T.~R. Martinez, Estimating the potential for combining learning
  models, in: Proceedings of the ICML Workshop on Meta-Learning, 2005, pp.
  68--75.

\bibitem{Dietterich2000_MLJ}
T.~G. Dietterich, An experimental comparison of three methods for constructing
  ensembles of decision trees: Bagging, boosting, and randomization, Machine
  Learning 40~(2) (2000) 139--157.

\bibitem{McDonald2003}
R.~A. McDonald, D.~J. Hand, I.~A. Eckley, An empirical comparison of three
  boosting algorithms on real data sets with artificial class noise, in:
  Proceedings of the 4th International workshop on Multiple Classifier Systems,
  2003, pp. 35--44.

\bibitem{Melville2004}
P.~Melville, N.~Shah, L.~Mihalkova, R.~J. Mooney, Experiments on ensembles with
  missing and noisy data., in: Multiple Classifier Systems, Vol. 3077 of
  Lecture Notes in Computer Science, 2004, pp. 293--302.

\bibitem{Ng+Jordan:2001}
A.~Y. Ng, M.~I. Jordan, On discriminative vs. generative classifiers: A
  comparison of logistic regression and naive bayes, in: Advances in Neural
  Information Processing Systems 14, 2001, pp. 841--848.

\bibitem{Lee2011}
J.~Lee, C.~Giraud-Carrier, A metric for unsupervised metalearning, Intelligent
  Data Analysis 15~(6) (2011) 827--841.

\bibitem{weka2009}
M.~Hall, E.~Frank, G.~Holmes, B.~Pfahringer, P.~Reutemann, I.~H. Witten, The
  weka data mining software: an update, SIGKDD Explorations Newsletter 11~(1)
  (2009) 10--18.

\bibitem{UCI2010}
A.~Frank, A.~Asuncion, \href{http://archive.ics.uci.edu/ml}{{UCI} machine
  learning repository} (2010).
\newline\urlprefix\url{http://archive.ics.uci.edu/ml}

\bibitem{Demsar2006}
J.~Dem\v{s}ar, Statistical comparisons of classifiers over multiple data sets,
  Journal of Machine Learning Research 7 (2006) 1--30.

\bibitem{Gamberger1999}
D.~Gamberger, N.~Lavrac, C.~Groselj, Experiments with noise filtering in a
  medical domain, in: Proceedings of the 16th International Conference on
  Machine Learning, 1999, pp. 143--151.

\bibitem{Brodley1995}
C.~E. Brodley, P.~E. Utgoff, Multivariate decision trees, Machine Learning
  19~(1) (1995) 45--77.

\bibitem{Khoshgoftaar2007}
T.~Khoshgoftaar, P.~Rebours, Improving software quality prediction by noise
  filtering techniques, Journal of Computer Science and Technology 22 (2007)
  387--396.

\bibitem{Kass1995}
R.~E. Kass, L.~Wassermann, {A reference Bayesian test for nested hypotheses and
  its relationship to the Schwarz criterion}, Journal of the American
  Statistical Association 90~(431) (1995) 928--934.

\bibitem{KEEL}
J.~Alcal\'{a}-Fdez, L.~S\'{a}nchez, S.~Garc\'{i}a, M.~J.~D. Jesus, S.~Ventura,
  J.~M. Garrell, J.~Otero, J.~Bacardit, V.~M. Rivas, J.~C. Fern\'{a}ndez,
  F.~Herrera, Keel: A software tool to assess evolutionary algorithms for data
  mining problems, Soft Computing 13~(3) (2008) 307--318.

\bibitem{Smith2012_IH}
M.~R. Smith, T.~Martinez, C.~Giraud-Carrier, An instance level analysis of data
  complexity, Machine Learning (2013) in press, 32 pages.

\end{thebibliography}







\appendix

\section{Significance Table for No Artificial Noise}
\label{section:sigTables}
This section provides Table \ref{table:LAsig} that compares the investigated weighting and filtering techniques with no noise handling for the investigated learning algorithms (C4.5, IB5, MLP, random forest, and RIPPER).
Table \ref{table:LAsig} provides more information for the data shown in Figure 5.

\begin{center}
\begin{longtable}{l|ccccc}
\multicolumn{6}{p{\textwidth}}{Table \thetable{}: A comparison of the investigated noise handling approaches averaged over all of 54 data sets 
for all of the investigated learning algorithms.
The underlined \textit{p}-values in the gray cells represent cases where noise handling significantly increases the accuracy; the bold \textit{p}-values represent cases where noise handling significantly decreases the accuracy.}
\label{table:LAsig} \\

\endfirsthead

\multicolumn{6}{p{\textwidth}}{Table \thetable{} (cont): A comparison of the investigated noise handling approaches across the various noise levels for all of the investigated learning algorithms. }\\
\multicolumn{6}{p{\textwidth}}{}\\
noise & 0\% & 10\% & 20\% & 30\% & 40\% \\
\hline
\endhead

\hline \multicolumn{6}{c}{Continued on next page}\\ \hline
\endfoot

\hline
\endlastfoot

\multicolumn{6}{c}{C4dot5}\\
noise & 0\% & 10\% & 20\% & 30\% & 40\% \\
\hline
orig & 79.31 & 77.07 & 74.47 & 69.89 & 63.31 \\
\hline
NICD: $\mathcal{L}$-Weighted & 78.36 & 77.39 & 76.44 & 74.32 & 70.84 \\
\textit{p}-val & 0.475 & \cellcolor{gray}\underline{0.954} & \cellcolor{gray}\underline{1.000} & \cellcolor{gray}\underline{1.000} & \cellcolor{gray}\underline{1.000} \\
Count & 26,1,27 & 20,1,33 & 13,0,41 & 11,0,43 & 10,0,43 \\
\hline
Biased-Weighted & 79.29 & 77.22 & 75.28 & 71.06 & 65.74 \\
\textit{p}-val & 0.658 & 0.845 & \cellcolor{gray}\underline{1.000} & \cellcolor{gray}\underline{1.000} & \cellcolor{gray}\underline{1.000} \\
Count & 23,7,24 & 20,4,30 & 11,4,39 & 13,7,34 & 12,1,41 \\
\hline
PWEM & 76.41 & 74.50 & 73.34 & 70.94 & 68.28 \\
\textit{p}-val & 0.062 & 0.059 & 0.687 & 0.931 & \cellcolor{gray}\underline{1.000} \\
Count & 30,3,21 & 31,0,23 & 22,2,30 & 22,0,32 & 15,0,39 \\
\hline
NICD: $\mathcal{L}$-Filter & 79.55 & 78.35 & 76.65 & 73.42 & 69.14 \\
\textit{p}-val & 0.913 & \cellcolor{gray}\underline{1.000} & \cellcolor{gray}\underline{1.000} & \cellcolor{gray}\underline{1.000} & \cellcolor{gray}\underline{1.000} \\
Count & 18,11,25 & 12,1,41 & 5,1,48 & 5,2,47 & 6,1,47 \\
\hline
Biased-Filter & 79.34 & 77.49 & 75.17 & 71.32 & 65.58 \\
\textit{p}-val & 0.422 & 0.923 & \cellcolor{gray}\underline{0.995} & \cellcolor{gray}\underline{1.000} & \cellcolor{gray}\underline{1.000} \\
Count & 25,7,22 & 22,4,28 & 15,4,35 & 14,3,37 & 12,2,40 \\
\hline
RENN & 76.83 & 76.11 & 74.71 & 71.95 & 66.29 \\
\textit{p}-val & \textbf{0.002} & 0.183 & 0.927 & \cellcolor{gray}\underline{0.995} & \cellcolor{gray}\underline{0.992} \\
Count & 32,3,19 & 28,0,26 & 21,0,33 & 17,3,34 & 20,0,34 \\
\hline
ANR & 67.25 & 65.97 & 64.12 & 62.39 & 59.12 \\
\textit{p}-val & $\mathbf{<0.001}$ & $\mathbf{<0.001}$ & \textbf{0.001} & 0.072 & 0.535 \\
Count & 34,3,17 & 32,2,20 & 33,0,21 & 29,1,24 & 25,1,28 \\
\hline
ClassificationFilter & 73.65 & 72.11 & 69.63 & 66.54 & 60.61 \\
\textit{p}-val & \textbf{0.009} & 0.139 & 0.067 & 0.223 & 0.382 \\
Count & 32,3,19 & 27,2,25 & 29,1,24 & 32,1,21 & 24,0,30 \\
\hline
CVCommitteesFilter & 79.41 & 77.35 & 74.60 & 70.11 & 63.64 \\
\textit{p}-val & 0.935 & \cellcolor{gray}\underline{0.979} & 0.838 & 0.929 & 0.878 \\
Count & 18,10,26 & 17,5,32 & 19,5,30 & 21,4,29 & 20,6,28 \\
\hline
EnsembleFilter & 79.53 & 77.88 & 76.15 & 73.62 & 68.66 \\
\textit{p}-val & 0.940 & \cellcolor{gray}\underline{0.974} & \cellcolor{gray}\underline{1.000} & \cellcolor{gray}\underline{1.000} & \cellcolor{gray}\underline{1.000} \\
Count & 24,2,28 & 19,4,31 & 10,2,42 & 11,0,43 & 9,1,44 \\
\hline
IterPartitionFilter & 79.31 & 77.33 & 74.82 & 70.41 & 64.31 \\
\textit{p}-val & 0.782 & \cellcolor{gray}\underline{0.955} & 0.928 & \cellcolor{gray}\underline{0.954} & \cellcolor{gray}\underline{0.969} \\
Count & 16,13,25 & 18,6,30 & 19,3,32 & 18,4,32 & 20,3,31 \\
\hline
SaturationFilter & 78.70 & 77.06 & 74.37 & 70.50 & 64.56 \\
\textit{p}-val & 0.418 & 0.336 & 0.624 & 0.872 & 0.907 \\
Count & 21,9,20 & 23,6,22 & 21,6,24 & 17,5,29 & 21,5,25 \\
\hline
\multicolumn{6}{c}{}\\
\multicolumn{6}{c}{IB5}\\
noise & 0\% & 10\% & 20\% & 30\% & 40\% \\
\hline
orig & 79.37 & 77.93 & 74.76 & 70.32 & 62.94 \\
\hline
NICD: $\mathcal{L}$-Weighted & 78.72 & 78.09 & 76.77 & 74.80 & 70.30 \\
\textit{p}-val & 0.162 & \cellcolor{gray}\underline{0.954} & \cellcolor{gray}\underline{1.000} & \cellcolor{gray}\underline{1.000} & \cellcolor{gray}\underline{1.000} \\
Count & 27,4,23 & 16,3,35 & 12,0,42 & 9,1,44 & 6,0,47 \\
\hline
Biased-Weighted & 78.34 & 77.41 & 75.39 & 71.59 & 64.92 \\
\textit{p}-val & $\mathbf{<0.001}$ & 0.179 & \cellcolor{gray}\underline{0.996} & \cellcolor{gray}\underline{1.000} & \cellcolor{gray}\underline{1.000} \\
Count & 32,7,15 & 22,8,24 & 14,2,38 & 11,1,42 & 8,3,43 \\
\hline
PWEM & 78.02 & 77.54 & 76.12 & 73.56 & 67.18 \\
\textit{p}-val & \textbf{0.003} & 0.605 & \cellcolor{gray}\underline{1.000} & \cellcolor{gray}\underline{1.000} & \cellcolor{gray}\underline{1.000} \\
Count & 33,3,18 & 23,0,31 & 16,1,37 & 11,0,43 & 7,1,46 \\
\hline
NICD: $\mathcal{L}$-Filter & 79.40 & 78.35 & 76.62 & 74.38 & 69.67 \\
\textit{p}-val & 0.854 & \cellcolor{gray}\underline{0.999} & \cellcolor{gray}\underline{1.000} & \cellcolor{gray}\underline{1.000} & \cellcolor{gray}\underline{1.000} \\
Count & 22,9,23 & 15,1,38 & 11,1,42 & 10,0,44 & 7,3,44 \\
\hline
Biased-Filter & 76.99 & 75.98 & 74.33 & 71.82 & 66.12 \\
\textit{p}-val & $\mathbf{<0.001}$ & \textbf{0.003} & 0.752 & \cellcolor{gray}\underline{0.981} & \cellcolor{gray}\underline{0.999} \\
Count & 35,4,15 & 33,2,19 & 22,0,32 & 18,0,36 & 13,0,41 \\
\hline
RENN & 76.99 & 75.98 & 74.33 & 71.82 & 66.12 \\
\textit{p}-val & $\mathbf{<0.001}$ & \textbf{0.003} & 0.752 & \cellcolor{gray}\underline{0.981} & \cellcolor{gray}\underline{0.999} \\
Count & 35,4,15 & 33,2,19 & 22,0,32 & 18,0,36 & 13,0,41 \\
\hline
ANR & 67.64 & 66.48 & 65.17 & 63.21 & 60.03 \\
\textit{p}-val & $\mathbf{<0.001}$ & $\mathbf{<0.001}$ & \textbf{0.047} & 0.261 & 0.805 \\
Count & 34,7,13 & 34,0,20 & 27,0,27 & 27,0,27 & 20,1,33 \\
\hline
ClassificationFilter & 73.09 & 71.81 & 69.30 & 66.91 & 61.23 \\
\textit{p}-val & $\mathbf{<0.001}$ & \textbf{0.009} & 0.116 & 0.599 & 0.791 \\
Count & 36,4,14 & 29,3,22 & 26,0,28 & 22,0,32 & 17,0,37 \\
\hline
CVCommitteesFilter & 79.37 & 78.38 & 76.50 & 73.21 & 66.36 \\
\textit{p}-val & 0.492 & \cellcolor{gray}\underline{0.997} & \cellcolor{gray}\underline{1.000} & \cellcolor{gray}\underline{1.000} & \cellcolor{gray}\underline{1.000} \\
Count & 22,14,18 & 17,3,34 & 9,2,43 & 9,4,41 & 5,4,45 \\
\hline
EnsembleFilter & 79.06 & 77.99 & 76.33 & 73.51 & 68.92 \\
\textit{p}-val & 0.140 & 0.931 & \cellcolor{gray}\underline{1.000} & \cellcolor{gray}\underline{1.000} & \cellcolor{gray}\underline{1.000} \\
Count & 32,1,21 & 20,3,31 & 14,2,38 & 15,0,39 & 9,2,43 \\
\hline
IterPartitionFilter & 79.27 & 78.20 & 76.33 & 73.63 & 67.90 \\
\textit{p}-val & 0.465 & \cellcolor{gray}\underline{0.984} & \cellcolor{gray}\underline{1.000} & \cellcolor{gray}\underline{1.000} & \cellcolor{gray}\underline{1.000} \\
Count & 23,10,21 & 17,4,33 & 11,4,39 & 8,4,42 & 10,2,42 \\
\hline
SaturationFilter & 79.08 & 77.28 & 74.84 & 70.43 & 63.74 \\
\textit{p}-val & 0.206 & 0.530 & \cellcolor{gray}\underline{0.992} & \cellcolor{gray}\underline{0.968} & \cellcolor{gray}\underline{0.998} \\
Count & 24,9,18 & 19,5,27 & 12,5,34 & 15,4,32 & 13,5,33 \\
\hline
\multicolumn{6}{c}{}\\
\multicolumn{6}{c}{MLP}\\
noise & 0\% & 10\% & 20\% & 30\% & 40\% \\
\hline
orig & 81.67 & 77.53 & 72.41 & 67.41 & 60.81 \\
\hline
NICD: $\mathcal{L}$-Weighted & 82.26 & 80.71 & 78.64 & 75.33 & 69.12 \\
\textit{p}-val & \cellcolor{gray}\underline{0.974} & \cellcolor{gray}\underline{1.000} & \cellcolor{gray}\underline{1.000} & \cellcolor{gray}\underline{1.000} & \cellcolor{gray}\underline{1.000} \\
Count & 18,3,33 & 9,1,44 & 6,0,48 & 6,1,47 & 3,0,50 \\
\hline
Biased-Weighted & 81.49 & 78.19 & 73.84 & 69.14 & 62.10 \\
\textit{p}-val & 0.718 & \cellcolor{gray}\underline{1.000} & \cellcolor{gray}\underline{1.000} & \cellcolor{gray}\underline{1.000} & \cellcolor{gray}\underline{1.000} \\
Count & 23,5,26 & 11,1,42 & 13,0,41 & 14,0,40 & 15,0,39 \\
\hline
PWEM & 82.79 & 79.67 & 76.42 & 71.90 & 65.95 \\
\textit{p}-val & 0.737 & \cellcolor{gray}\underline{1.000} & \cellcolor{gray}\underline{1.000} & \cellcolor{gray}\underline{1.000} & \cellcolor{gray}\underline{1.000} \\
Count & 23,3,28 & 13,0,41 & 10,1,43 & 8,1,45 & 7,1,46 \\
\hline
NICD: $\mathcal{L}$-Filter & 81.80 & 80.66 & 78.17 & 74.82 & 69.44 \\
\textit{p}-val & 0.584 & \cellcolor{gray}\underline{1.000} & \cellcolor{gray}\underline{1.000} & \cellcolor{gray}\underline{1.000} & \cellcolor{gray}\underline{1.000} \\
Count & 27,4,23 & 8,0,46 & 8,1,45 & 5,1,48 & 4,0,50 \\
\hline
Biased-Filter & 81.39 & 78.16 & 75.04 & 70.57 & 64.35 \\
\textit{p}-val & 0.339 & \cellcolor{gray}\underline{0.997} & \cellcolor{gray}\underline{1.000} & \cellcolor{gray}\underline{1.000} & \cellcolor{gray}\underline{1.000} \\
Count & 24,10,20 & 15,7,32 & 12,1,41 & 10,1,43 & 10,2,42 \\
\hline
RENN & 78.80 & 77.80 & 76.19 & 73.30 & 67.43 \\
\textit{p}-val & $\mathbf{<0.001}$ & 0.790 & \cellcolor{gray}\underline{0.999} & \cellcolor{gray}\underline{1.000} & \cellcolor{gray}\underline{1.000} \\
Count & 38,1,15 & 25,1,28 & 20,0,34 & 15,1,38 & 15,0,39 \\
\hline
ANR & 69.17 & 66.99 & 64.92 & 62.31 & 58.93 \\
\textit{p}-val & $\mathbf{<0.001}$ & \textbf{0.004} & 0.178 & 0.425 & 0.883 \\
Count & 37,2,15 & 33,0,21 & 28,0,26 & 24,0,30 & 21,0,33 \\
\hline
ClassificationFilter & 75.65 & 73.83 & 71.22 & 67.51 & 61.22 \\
\textit{p}-val & \textbf{0.002} & 0.545 & 0.937 & \cellcolor{gray}\underline{0.988} & \cellcolor{gray}\underline{0.963} \\
Count & 34,2,18 & 26,0,28 & 21,1,32 & 18,0,36 & 20,0,34 \\
\hline
CVCommitteesFilter & 82.33 & 79.83 & 76.68 & 71.50 & 65.02 \\
\textit{p}-val & 0.947 & \cellcolor{gray}\underline{1.000} & \cellcolor{gray}\underline{1.000} & \cellcolor{gray}\underline{1.000} & \cellcolor{gray}\underline{1.000} \\
Count & 20,4,30 & 11,2,41 & 10,2,42 & 11,2,41 & 15,0,39 \\
\hline
EnsembleFilter & 81.62 & 80.02 & 77.44 & 73.85 & 68.04 \\
\textit{p}-val & 0.504 & \cellcolor{gray}\underline{1.000} & \cellcolor{gray}\underline{1.000} & \cellcolor{gray}\underline{1.000} & \cellcolor{gray}\underline{1.000} \\
Count & 28,1,25 & 15,0,39 & 11,1,42 & 10,1,43 & 9,0,45 \\
\hline
IterPartitionFilter & 82.18 & 80.08 & 77.31 & 73.02 & 67.02 \\
\textit{p}-val & 0.862 & \cellcolor{gray}\underline{1.000} & \cellcolor{gray}\underline{1.000} & \cellcolor{gray}\underline{1.000} & \cellcolor{gray}\underline{1.000} \\
Count & 21,4,29 & 12,1,41 & 11,1,42 & 9,3,42 & 10,2,42 \\
\hline
SaturationFilter & 81.07 & 78.15 & 74.89 & 69.47 & 63.20 \\
\textit{p}-val & 0.216 & \cellcolor{gray}\underline{0.987} & \cellcolor{gray}\underline{0.997} & \cellcolor{gray}\underline{0.992} & \cellcolor{gray}\underline{0.996} \\
Count & 26,4,20 & 13,3,34 & 18,4,29 & 16,3,32 & 16,2,33 \\
\hline
\multicolumn{6}{c}{}\\
\multicolumn{6}{c}{RandForest}\\
noise & 0\% & 10\% & 20\% & 30\% & 40\% \\
\hline
orig & 81.18 & 77.97 & 72.93 & 67.07 & 59.81 \\
\hline
NICD: $\mathcal{L}$-Weighted & 80.82 & 79.73 & 78.08 & 75.87 & 70.70 \\
\textit{p}-val & 0.323 & \cellcolor{gray}\underline{1.000} & \cellcolor{gray}\underline{1.000} & \cellcolor{gray}\underline{1.000} & \cellcolor{gray}\underline{1.000} \\
Count & 28,2,24 & 11,0,43 & 7,0,47 & 2,0,51 & 3,1,49 \\
\hline
Biased-Weighted & 80.94 & 78.24 & 74.01 & 68.25 & 60.93 \\
\textit{p}-val & 0.281 & \cellcolor{gray}\underline{0.965} & \cellcolor{gray}\underline{1.000} & \cellcolor{gray}\underline{1.000} & \cellcolor{gray}\underline{1.000} \\
Count & 26,4,24 & 22,0,32 & 8,0,46 & 13,1,40 & 11,0,43 \\
\hline
PWEM & 81.51 & 78.73 & 76.37 & 72.54 & 66.03 \\
\textit{p}-val & 0.125 & \cellcolor{gray}\underline{0.956} & \cellcolor{gray}\underline{1.000} & \cellcolor{gray}\underline{1.000} & \cellcolor{gray}\underline{1.000} \\
Count & 34,1,19 & 21,1,32 & 9,2,43 & 5,0,49 & 6,0,48 \\
\hline
NICD: $\mathcal{L}$-Filter & 81.66 & 79.91 & 78.23 & 75.45 & 70.11 \\
\textit{p}-val & 0.873 & \cellcolor{gray}\underline{1.000} & \cellcolor{gray}\underline{1.000} & \cellcolor{gray}\underline{1.000} & \cellcolor{gray}\underline{1.000} \\
Count & 24,2,28 & 9,1,44 & 5,0,49 & 2,0,52 & 2,1,51 \\
\hline
Biased-Filter & 81.16 & 78.13 & 74.25 & 69.32 & 62.80 \\
\textit{p}-val & 0.584 & 0.949 & \cellcolor{gray}\underline{1.000} & \cellcolor{gray}\underline{1.000} & \cellcolor{gray}\underline{1.000} \\
Count & 21,12,21 & 18,5,31 & 5,7,42 & 11,4,39 & 14,6,34 \\
\hline
RENN & 78.20 & 77.25 & 75.62 & 73.32 & 67.52 \\
\textit{p}-val & $\mathbf{<0.001}$ & 0.679 & \cellcolor{gray}\underline{0.999} & \cellcolor{gray}\underline{1.000} & \cellcolor{gray}\underline{1.000} \\
Count & 35,2,17 & 25,0,29 & 17,0,37 & 8,2,44 & 13,0,41 \\
\hline
ANR & 68.20 & 66.76 & 65.09 & 62.76 & 59.27 \\
\textit{p}-val & $\mathbf{<0.001}$ & $\mathbf{<0.001}$ & 0.123 & 0.708 & \cellcolor{gray}\underline{0.965} \\
Count & 36,3,15 & 30,0,24 & 25,1,28 & 22,1,31 & 19,1,34 \\
\hline
ClassificationFilter & 75.11 & 73.18 & 70.44 & 67.58 & 60.75 \\
\textit{p}-val & \textbf{0.012} & 0.268 & 0.829 & \cellcolor{gray}\underline{0.996} & \cellcolor{gray}\underline{0.993} \\
Count & 32,1,21 & 22,0,32 & 18,2,34 & 15,0,39 & 17,0,37 \\
\hline
CVCommitteesFilter & 81.72 & 79.53 & 76.88 & 72.40 & 64.97 \\
\textit{p}-val & \cellcolor{gray}\underline{0.958} & \cellcolor{gray}\underline{1.000} & \cellcolor{gray}\underline{1.000} & \cellcolor{gray}\underline{1.000} & \cellcolor{gray}\underline{1.000} \\
Count & 21,2,31 & 13,1,40 & 6,0,48 & 3,1,50 & 4,1,49 \\
\hline
EnsembleFilter & 80.91 & 79.52 & 77.32 & 74.01 & 68.86 \\
\textit{p}-val & 0.253 & \cellcolor{gray}\underline{0.999} & \cellcolor{gray}\underline{1.000} & \cellcolor{gray}\underline{1.000} & \cellcolor{gray}\underline{1.000} \\
Count & 31,1,22 & 14,1,39 & 9,0,45 & 8,0,46 & 5,2,47 \\
\hline
IterPartitionFilter & 81.51 & 79.75 & 77.61 & 74.02 & 67.40 \\
\textit{p}-val & 0.777 & \cellcolor{gray}\underline{1.000} & \cellcolor{gray}\underline{1.000} & \cellcolor{gray}\underline{1.000} & \cellcolor{gray}\underline{1.000} \\
Count & 24,4,26 & 9,2,43 & 6,0,48 & 2,1,51 & 4,0,50 \\
\hline
SaturationFilter & 80.28 & 78.16 & 74.56 & 69.46 & 62.03 \\
\textit{p}-val & 0.089 & \cellcolor{gray}\underline{0.954} & \cellcolor{gray}\underline{0.999} & \cellcolor{gray}\underline{1.000} & \cellcolor{gray}\underline{0.998} \\
Count & 29,2,19 & 17,0,34 & 13,1,37 & 10,2,39 & 17,0,34 \\
\hline
\multicolumn{6}{c}{}\\
\multicolumn{6}{c}{RIPPER}\\
noise & 0\% & 10\% & 20\% & 30\% & 40\% \\
\hline
orig & 78.35 & 76.41 & 73.54 & 69.94 & 65.07 \\
\hline
NICD: $\mathcal{L}$-Weighted & 77.86 & 76.62 & 75.53 & 73.38 & 69.53 \\
\textit{p}-val & 0.240 & 0.861 & \cellcolor{gray}\underline{1.000} & \cellcolor{gray}\underline{1.000} & \cellcolor{gray}\underline{1.000} \\
Count & 26,2,26 & 20,3,31 & 7,1,46 & 5,0,48 & 9,0,44 \\
\hline
Biased-Weighted & 77.98 & 76.28 & 74.50 & 70.70 & 65.97 \\
\textit{p}-val & 0.146 & 0.472 & \cellcolor{gray}\underline{1.000} & \cellcolor{gray}\underline{0.985} & 0.936 \\
Count & 29,4,21 & 24,1,29 & 15,2,37 & 20,2,32 & 19,2,33 \\
\hline
PWEM & 74.17 & 71.94 & 70.68 & 68.57 & 64.82 \\
\textit{p}-val & $\mathbf{<0.001}$ & $\mathbf{<0.001}$ & \textbf{0.013} & 0.445 & 0.746 \\
Count & 39,1,14 & 37,3,14 & 34,1,19 & 28,0,26 & 23,0,31 \\
\hline
NICD: $\mathcal{L}$-Filter & 78.98 & 77.82 & 76.43 & 73.97 & 69.90 \\
\textit{p}-val & 0.893 & \cellcolor{gray}\underline{1.000} & \cellcolor{gray}\underline{1.000} & \cellcolor{gray}\underline{1.000} & \cellcolor{gray}\underline{1.000} \\
Count & 22,5,27 & 10,3,41 & 7,1,46 & 7,1,46 & 9,0,45 \\
\hline
Biased-Filter & 77.20 & 75.70 & 73.48 & 70.39 & 65.78 \\
\textit{p}-val & \textbf{0.025} & 0.293 & \cellcolor{gray}\underline{0.950} & \cellcolor{gray}\underline{0.975} & \cellcolor{gray}\underline{0.988} \\
Count & 30,7,17 & 24,7,23 & 17,9,28 & 14,13,27 & 16,9,29 \\
\hline
RENN & 76.65 & 75.88 & 74.59 & 71.81 & 66.90 \\
\textit{p}-val & \textbf{0.004} & 0.480 & \cellcolor{gray}\underline{0.979} & \cellcolor{gray}\underline{0.981} & 0.946 \\
Count & 34,2,18 & 24,5,25 & 21,0,33 & 18,1,35 & 23,0,31 \\
\hline
ANR & 68.18 & 66.89 & 64.12 & 62.59 & 59.78 \\
\textit{p}-val & $\mathbf{<0.001}$ & $\mathbf{<0.001}$ & \textbf{0.002} & 0.072 & 0.170 \\
Count & 40,1,12 & 33,2,18 & 33,1,20 & 25,2,27 & 24,1,29 \\
\hline
ClassificationFilter & 72.48 & 70.94 & 68.96 & 66.65 & 61.38 \\
\textit{p}-val & \textbf{0.001} & \textbf{0.009} & \textbf{0.047} & 0.355 & 0.077 \\
Count & 34,3,17 & 35,0,19 & 30,1,23 & 24,1,29 & 30,1,23 \\
\hline
CVCommitteesFilter & 78.83 & 77.19 & 74.85 & 71.15 & 67.24 \\
\textit{p}-val & \cellcolor{gray}\underline{0.965} & \cellcolor{gray}\underline{0.995} & \cellcolor{gray}\underline{1.000} & \cellcolor{gray}\underline{0.999} & \cellcolor{gray}\underline{1.000} \\
Count & 16,8,30 & 16,6,32 & 13,5,36 & 12,2,40 & 14,4,36 \\
\hline
EnsembleFilter & 78.87 & 77.29 & 75.90 & 73.19 & 69.22 \\
\textit{p}-val & 0.875 & \cellcolor{gray}\underline{0.998} & \cellcolor{gray}\underline{1.000} & \cellcolor{gray}\underline{1.000} & \cellcolor{gray}\underline{1.000} \\
Count & 23,3,28 & 15,4,35 & 7,3,44 & 3,2,49 & 7,0,47 \\
\hline
IterPartitionFilter & 78.78 & 77.23 & 75.41 & 72.36 & 67.87 \\
\textit{p}-val & 0.744 & \cellcolor{gray}\underline{0.986} & \cellcolor{gray}\underline{1.000} & \cellcolor{gray}\underline{1.000} & \cellcolor{gray}\underline{0.997} \\
Count & 23,5,26 & 16,4,34 & 15,1,38 & 15,2,37 & 18,2,34 \\
\hline
SaturationFilter & 78.46 & 76.49 & 74.42 & 70.82 & 64.66 \\
\textit{p}-val & 0.402 & 0.116 & 0.856 & \cellcolor{gray}\underline{0.952} & 0.132 \\
Count & 22,7,21 & 28,3,20 & 19,4,28 & 18,3,30 & 26,4,21 \\

\end{longtable}

\end{center}

\section{Comparison of 3-Ensemble with Noise Handling}
\label{appendix_3-ens}
This section provides Table \ref{table:voting_3} that compares the 3-ensemble with $\mathcal{L}$-weighting and $\mathcal{L}$-filtering for the investigated learning algorithms (C4.5, IB5, MLP, random forest, and RIPPER).
$\mathcal{L}$-weighting and $\mathcal{L}$-filtering were chosen as representative noise handling techniques since they obtained the most significant and highest increase in classification accuracy.
The MLP trained with backpropagation achieves the highest classification accuracy without any artificial noise added.

\begin{center}
\setlength{\tabcolsep}{1.4pt}
\begin{longtable}{l|ccccc||ccccc}
\multicolumn{11}{p{\textwidth}}{Table \thetable{}: A comparison of the 3-ensemble against $\mathcal{L}$-weighting and $\mathcal{L}$-filtering for the 5 considered learning algorithms averaged over the 54 data sets.
Bold \textit{p}-values represent cases where the 3-ensemble achieves significantly higher accuracy; the underlined, gray cells represent cases where $\mathcal{L}$-weighting or $\mathcal{L}$-filtering achieves significantly higher accuracy.}
\label{table:voting_3} \\

\endfirsthead

\multicolumn{11}{p{\textwidth}}{Table \thetable{} (cont): A comparison of the 3-ensemble against $\mathcal{L}$-weighting and $\mathcal{L}$-filtering for the 5 considered learning algorithms. }\\
\multicolumn{11}{p{\textwidth}}{}\\
noise & 0\% & 10\% & 20\% & 30\% & 40\% & 0\% & 10\% & 20\% & 30\% & 40\%  \\
\hline
\endhead

\hline \multicolumn{11}{c}{Continued on next page}\\ \hline
\endfoot

\cline{1-6}
\endlastfoot

noise  & 0\% & 10\% & 20\% & 30\% & 40\% & 0\% & 10\% & 20\% & 30\% & 40\% \\
\hline
3-Ens & 82.64 & 80.75 & 77.14 & 72.37 & 64.94 &  \\
\hline
& \multicolumn{5}{c||}{C4.5}& \multicolumn{5}{c}{IB5} \\
\hline
$\mathcal{L}$-Weight & 78.19 & 77.03 & 76.03 & 74.12 & 70.88 & 78.72 & 77.86 & 77.05 & 75.11 & 70.66 \\
\textit{p}-val & $\textbf{0.000}$ & $\textbf{0.000}$ & 0.281 & \cellcolor{gray}\underline{0.961} & \cellcolor{gray}\underline{1.000}& $\textbf{0.000}$ & $\textbf{0.000}$ & 0.785 & \cellcolor{gray}\underline{1.000}& \cellcolor{gray}\underline{1.000}\\
Count & \begin{footnotesize}40,1,13\end{footnotesize} & \begin{footnotesize}37,0,16\end{footnotesize} & \begin{footnotesize}29,0,24\end{footnotesize} & \begin{footnotesize}21,1,32\end{footnotesize} & \begin{footnotesize}11,0,43\end{footnotesize} & \begin{footnotesize}43,1,10\end{footnotesize} & \begin{footnotesize}30,3,20\end{footnotesize} & \begin{footnotesize}23,0,31\end{footnotesize} & \begin{footnotesize}14,0,40\end{footnotesize} & \begin{footnotesize}9,0,45\end{footnotesize} \\
\hline
$\mathcal{L}$-Filter & 79.55 & 78.35 & 76.65 & 73.42 & 69.14 & 79.4 & 78.35 & 76.62 & 74.38 & 69.67 \\
\textit{p}-val & $\textbf{0.000}$ & $\textbf{0.000}$ & 0.393 & 0.911 & \cellcolor{gray}\underline{1.000}& $\textbf{0.000}$ & $\textbf{0.000}$ & 0.556 & \cellcolor{gray}\underline{0.998} & \cellcolor{gray}\underline{1.000}\\
Count & \begin{footnotesize}41,2,11\end{footnotesize} & \begin{footnotesize}37,0,17\end{footnotesize} & \begin{footnotesize}28,0,26\end{footnotesize} & \begin{footnotesize}23,1,30\end{footnotesize} & \begin{footnotesize}13,0,41\end{footnotesize} & \begin{footnotesize}42,1,11\end{footnotesize} & \begin{footnotesize}37,3,14\end{footnotesize} & \begin{footnotesize}25,0,29\end{footnotesize} & \begin{footnotesize}18,0,36\end{footnotesize} & \begin{footnotesize}13,1,40\end{footnotesize} \\
\hline

\hline
& \multicolumn{5}{c||}{MLP}& \multicolumn{5}{c}{Random Forest} \\
\hline
$\mathcal{L}$-Weight & 82.26 & 80.70 & 78.44 & 75.16 & 69.31 & 80.82 & 79.72 & 78.28 & 76.12 & 71.01 \\
\textit{p}-val & 0.449 & 0.914 & \cellcolor{gray}\underline{1.000}& \cellcolor{gray}\underline{1.000}& \cellcolor{gray}\underline{1.000}& $\textbf{0.000}$ & 0.143 & \cellcolor{gray}\underline{0.996} & \cellcolor{gray}\underline{1.000}& \cellcolor{gray}\underline{1.000}\\
Count & \begin{footnotesize}23,6,25\end{footnotesize} & \begin{footnotesize}21,0,32\end{footnotesize} & \begin{footnotesize}12,0,42\end{footnotesize} & \begin{footnotesize}8,0,46\end{footnotesize} & \begin{footnotesize}8,0,46\end{footnotesize} & \begin{footnotesize}35,2,17\end{footnotesize} & \begin{footnotesize}27,1,25\end{footnotesize} & \begin{footnotesize}13,3,38\end{footnotesize} & \begin{footnotesize}12,0,42\end{footnotesize} & \begin{footnotesize}7,0,47\end{footnotesize} \\
\hline
$\mathcal{L}$-Filter & 81.80 & 80.66 & 78.17 & 74.82 & 69.44 & 81.66 & 79.91 & 78.23 & 75.45 & 70.11 \\
\textit{p}-val & \textbf{0.025} & 0.871 & \cellcolor{gray}\underline{0.997} & \cellcolor{gray}\underline{1.00}0& \cellcolor{gray}\underline{1.000}& \textbf{0.004} & 0.057 & \cellcolor{gray}\underline{0.997} & \cellcolor{gray}\underline{1.000}& \cellcolor{gray}\underline{1.000}\\
Count & \begin{footnotesize}32,2,20\end{footnotesize} & \begin{footnotesize}24,0,30\end{footnotesize} & \begin{footnotesize}16,0,38\end{footnotesize} & \begin{footnotesize}14,0,40\end{footnotesize} & \begin{footnotesize}10,0,44\end{footnotesize} & \begin{footnotesize}35,2,17\end{footnotesize} & \begin{footnotesize}31,1,22\end{footnotesize} & \begin{footnotesize}17,0,37\end{footnotesize} & \begin{footnotesize}11,1,42\end{footnotesize} & \begin{footnotesize}8,1,45\end{footnotesize} \\
\hline

& \multicolumn{5}{c||}{RIPPER}& \\
\cline{1-6}
$\mathcal{L}$-Weight & 77.86 & 76.54 & 75.47 & 73.45 & 69.63 \\
\textit{p}-val & $\textbf{0.000}$ & $\textbf{0.000}$ & 0.236 & 0.904 & \cellcolor{gray}\underline{1.000}\\
Count & \begin{footnotesize}39,3,12\end{footnotesize} & \begin{footnotesize}38,0,15\end{footnotesize} & \begin{footnotesize}30,1,22\end{footnotesize} & \begin{footnotesize}24,1,29\end{footnotesize} & \begin{footnotesize}15,1,38\end{footnotesize} \\
\cline{1-6}
$\mathcal{L}$-Filter & 78.98 & 77.82 & 76.43 & 73.97 & 69.9 \\
\textit{p}-val & $\textbf{0.000}$ & $\textbf{0.000}$ & 0.397 & \cellcolor{gray}\underline{0.977} & \cellcolor{gray}\underline{1.000}\\
Count & \begin{footnotesize}41,2,11\end{footnotesize} & \begin{footnotesize}38,0,16\end{footnotesize} & \begin{footnotesize}26,1,27\end{footnotesize} & \begin{footnotesize}20,1,33\end{footnotesize} & \begin{footnotesize}13,0,41\end{footnotesize} \\
\cline{1-6}
\end{longtable}
\end{center}

\end{document}